# Ensemble Methods for Multi-label Classification

Lior Rokach [1], Alon Schclar [2], Ehud Itach [1]

[1] Department of Information Systems Engineering, Ben-Gurion University of the Negev
P.O.B. 653, Beer-Sheva 84105, Israel
{liorrk, itach}@bgu.ac.il

[2] School of Computer Science, Academic College of Tel-Aviv Yafo
P.O.B 8401, Tel Aviv 61083, Israel
alonschc@mta.ac.il

**Abstract.** Ensemble methods have been shown to be an effective tool for solving multi-label classification tasks. In the RAndom $k$-labELsets (RAKEL) algorithm, each member of the ensemble is associated with a small randomly-selected subset of $k$ labels. Then, a single label classifier is trained according to each combination of elements in the subset. In this paper we adopt a similar approach, however, instead of randomly choosing subsets, we select the minimum required subsets of $k$ labels that cover all labels and meet additional constraints such as coverage of inter-label correlations. Construction of the cover is achieved by formulating the subset selection as a minimum set covering problem (SCP) and solving it by using approximation algorithms. Every cover needs only to be prepared once by offline algorithms. Once prepared, a cover may be applied to the classification of any given multi-label dataset whose properties conform with those of the cover. The contribution of this paper is two-fold. First, we introduce SCP as a general framework for constructing label covers while allowing the user to incorporate cover construction constraints. We demonstrate the effectiveness of this framework by proposing two construction constraints whose enforcement produces covers that improve the prediction performance of random selection. Second, we provide theoretical bounds that quantify the probabilities of random selection to produce covers that meet the proposed construction criteria. The experimental results indicate that the proposed methods improve multi-label classification accuracy and stability compared with the RAKEL algorithm and to other state-of-the-art algorithms.

## 1 Introduction

An inducer is an algorithm that constructs classifiers by learning a set of labeled examples (training set) whose classification (label value) is known *a-priori*. The classifier (also known as a classification model) can then be used to label unclassified instances. Commonly, the examples are associated with a single label which can assume either two (binary) or more (multiclass) distinct values. However, there are many cases where each instance needs to be associated with a number of labels. In these cases, there is a set of labels $L = \{\lambda_i\}$ that is associated with the training set. The classification of each training instance is given as values in a subset of labels taken from *L*. The subsets are not necessarily disjoint. For simplicity, it is usually assumed that every label is binary. Multi-label classification is employed in a wide range of applications such as text categorization [20, 47] (e.g., books associated with multiple genres) and medical diagnosis [8] (e.g., patients with multiple diseases) etc.

Tsoumakas and Katakis [41] divided multi-label classification methods into two main categories: *problem transformation* and *algorithm adaptation*. Problem transformation methods, which are the focus of this paper, transform the multi-label classification problem into several single-label classification problems while algorithm adaptation methods adjust single-label classifiers to handle multi-label data. The main weakness of methods that belong to the latter category is that they are mostly tailored for a specific classifier (e.g., SVM, decision tree), and thus lack generality. The first group, on the other hand, is more general and suits many cases and classifiers.

Two known multi-label classification methods that belong to the first group of methods are: *binary relevance* (BR) [5] and *label powerset* (LP) [4]. The BR method builds independent binary classifiers for each label ($\lambda_i$). Each classifier maps the original dataset to a single binary label with values $\lambda_i$, $\neg \lambda_i$. The classification of a new instance is given by the concatenation of the labels $\lambda_i$ that are produced by the classifiers. The major disadvantage of this approach is that it does not take into account inter-label correlations. Conversely, the LP method takes into account inter-label correlations by building a single classifier in which every unique combination of labels values constitutes a single label. The set of all combinations is also known as the powerset of $L$ and it is denoted by $P(L)$. One of the main drawbacks of this approach is that each combination may be associated with a very small number of instances since the number of all possible combinations is exponential in the number of labels.

In order to take into account inter-label correlations while avoiding the disadvantage of the LP method, it was suggested to construct an ensemble of single-label classifiers. A new instance is classified by integrating the outputs of the single-label classifiers. Tsoumakas and Vlahavas [45] and Tsoumakas *et al.* [43] presented the RAndom k-labELsets (RAKEL) algorithm - an effective ensemble method for solving multi-label classification tasks. Each ensemble member constructs an LP classifier based on a randomly chosen subset of $k$ labels. These subsets are referred to as *k-labelsets*. The classification of a new instance is achieved by thresholding the average of the binary decisions of each model for each label. The authors showed that RAKEL achieved high predictive performance compared to BR and LP methods. The simplicity of RAKEL together with its predictive performance made it the algorithm of choice for solving multi-label classification tasks. However, the random selection of subsets in RAKEL may negatively affect the ensemble's performance. We investigate this in Section 2. Specifically, the chosen subsets may not cover all labels and inter-label correlations. The importance of inter-label correlations can be demonstrated by the following scenario. Suppose that the behavior of two certain labels is similar in all of the instances of the training set e.g., whenever one label is associated with a certain instance, the other is associated with it as well and vice versa. Not considering the correlation between these labels may result in associating a test instance to only one of the labels and not the other. In Section 4.1 we provide analytical bounds on the coverage of inter-label correlation. Note that since each ensemble member is constructed as an LP classifier, high values of $k$ make this construction impractical due to the exponential number of enumerations required by the LP classifier. Accordingly, $k$ needs to be bounded by a moderate value – a guideline which is followed by the methods proposed in this paper.

The main challenge in the construction of an effective multi-label ensemble of classifiers that uses subset selection is to determine the label subsets for each ensemble member. Ideally, we would like to choose the minimal number of subsets that cover all labels and produce the best predictive performance. We denote the number of label subsets by $\sigma$. Two methods that consider inter-label correlations were previously proposed by us [24, 36]. Rather than randomly choosing the subsets [41], the subset selection problem was formulated as a set cov-ering problem (SCP) and an approximation algorithm was employed to derive the subsets. The SCP solution produced a compact ensemble that covered all possible labels using label sets of size $k$ while including all inter-label correlations between label subsets [36] of size $r < k$. The ensemble classifies a new sample by averaging the classification confidence of the base classifiers for each of the labels. The average is then compared to a threshold in order to derive the final decision for each label. The threshold value is tailored to the dataset in hand using a cross validation procedure.

In this paper we generalize our previous work [24, 36] and propose SCP as a general framework for the construction of multi-label ensemble classifiers. This formulation allows the application of a wide variety of optimization criteria to the subset construction process - thus, resulting in different covers of the original set of labels. We propose three algorithms for choosing the subsets. All three algorithms try to cover all labels by a minimal number of subsets or by a given number of label subsets. The algorithms differ by additional criteria they take into account. Specifically, each algorithm considers one or more of the following criteria: (a) label occurrence frequency in the ensemble; and (b) inter-label correlations coverage. The first algorithm chooses the subsets so that all labels will have an equal contribution to the ensemble. We refer to this algorithm as the BAlanced Label COntribution approach (BALCO). The second algorithm focuses on covering the highest possible number of inter-label correlations. This algorithm is referred to as the INter-LAbel Correlations algorithm (INLAC). The third algorithm combines the criteria of the first two algorithms i.e. it covers as much inter-label correlations as possible while maintaining an even number of label occurrences. We refer to this algorithm as the BALANced label contribution inter-label CORrelations (BALANCOR) algorithm. The INLAC and BALANCOR algorithms receive as input the level $r$ of inter-label correlation to cover e.g. correlations among label pairs ($r = 2$), triplets ($r = 3$), quartets ($r = 4$), etc. Additionally, we propose a data driven post processing algorithm (DD-BALANCOR) which maximizes the number of inter-label correlations that are taken into account. This is achieved by permuting the set of labels (which correspond to columns) of the constructed matrix according to the label dependencies of the dataset in hand. All four algorithms use a revised version of RAKEL (called RAKEL++) which averages the classifications confidence values (instead of using voting as in [41]) and employs a built-in procedure for deriving the best threshold value to the dataset in hand.

The rest of this paper is organized as follows: Section 2 describes the ensemble algorithm that is used for the various subset selection strategies. Section 3 describes the set cover problem and a greedy algorithm for approximating its solution. In Section 4, we show how the set cover problem can be used as a general framework for the construction of label subsets. We then propose three strategies for choosing the subsets by enforcing various constraints on the selection process. Experimental results are presented and discussed in Section 5. We conclude and suggest future research in Section 6.

## 2 The Ensemble Algorithm

### 2.1 Label subsets representation

In order to represent the label subsets that constitute each ensemble, we construct a binary matrix in which the columns correspond to the classes $L = \{\lambda_i\}$ and the each row corresponds to an ensemble member. The $i$-th row contains 1 in its $j$-th column if the label $\lambda_j$ is included in the subset of the $i$-th ensemble member. Table 1 il-lustrates a table which represents an ensemble of 8 members and 9 labels. For example, the sixth ensemble member is constructed based on the label subset $\{\lambda_1, \lambda_3, \lambda_6\}$.

**Table 1**: An example of a binary matrix that represents subsets of labels that are used to construct the ensemble members.

|   | $\lambda_1$ | $\lambda_2$ | $\lambda_3$ | $\lambda_4$ | $\lambda_5$ | $\lambda_6$ | $\lambda_7$ | $\lambda_8$ | $\lambda_9$ |
|---|---|---|---|---|---|---|---|---|---|
| $S_1$ | 1 | 0 | 0 | 0 | 0 | 1 | 1 | 1 | 0 |
| $S_2$ | 0 | 1 | 0 | 1 | 1 | 0 | 1 | 1 | 0 |
| $S_3$ | 1 | 0 | 1 | 0 | 0 | 1 | 0 | 1 | 0 |
| $S_4$ | 1 | 1 | 0 | 0 | 1 | 0 | 0 | 0 | 0 |
| $S_5$ | 0 | 1 | 0 | 1 | 1 | 0 | 1 | 1 | 0 |
| $S_6$ | 1 | 0 | 1 | 0 | 0 | 1 | 0 | 0 | 0 |
| $S_7$ | 0 | 1 | 0 | 1 | 1 | 1 | 1 | 0 | 0 |
| $S_8$ | 1 | 1 | 1 | 0 | 1 | 1 | 1 | 1 | 0 |

This representation is highly effective for deriving properties that can quantify the quality of the ensemble. For example, summing the rows can reveal the coverage level of each label where a zero sum indicates that the label is not covered by any of the subsets ($\lambda_9$ in Table 1). Furthermore, two columns that complement one another indicate that none of the ensemble members cover both of the labels associated with these columns. Thus any possible correlation between these labels is ignored by the ensemble (columns $\lambda_1$ and $\lambda_4$ in Table 1).

**2.2 Ensemble construction**

Given a matrix representation, each ensemble member is constructed as a multi-class classifier using the LP approach. Specifically, the *i*-th ensemble member is constructed based on the enumeration of all possible combinations of the selected classes (contain 1 in their corresponding column at the *i*-th row). Algorithm 1 presents the required steps for the construction of the ensemble. The input to the algorithm is a binary matrix *M* representing the subsets that were chosen according to a set of given criteria. Note that unlike the RAKEL algorithm, we separate between the subset selection and the ensemble construction. This allows the *offline* construction of optimal cover matrices which excludes the matrix construction from the complexity of the ensemble construction. This separation is advisable since finding an optimal set covering is NP-hard and even finding a sub-optimal solution might be computationally intensive.

---

**Algorithm 1: Construction of the ensemble classifier**

*Input:* *k*-labelset matrix – ***M***
      The training set - ***T***

*Output:* An ensemble of LP classifiers $h_i$

1. **for** $i \leftarrow 1$ to the number of rows in *M* **do**
2.     $Y_i \leftarrow$ the labelset represented by the *i*-th row in *M*;
3.     train an LP classifier $h_i : T \rightarrow P(Y_i)$ on *D*;
4. **end for**

---

Since each ensemble member constructs an LP classifier, large label subsets need to be avoided. Otherwise the number of training instances associated with each class is relatively small which makes it harder for the base-learning algorithm to differentiate among the classes. Accordingly, as in RAKEL, we are interested in binary constant weight codes in which all codewords share the same Hamming weight of *k* (i.e. the number of 1's is constant in all rows of the matrix). Due to this reason, one cannot simply use binary matrices that were con-structed for multi-class classification such as error-correcting output coding (ECOC) [13,19]. However, certain codes such as *Orthogonal array* [23] can be used for this purpose. An orthogonal array $OA(m, \sigma, d, t)$ is a matrix of $\sigma$ rows and $m$ columns, with every element being one of the $d$ values. The array has strength $t$ if, in every $t$ by $n$ submatrix, all the possible $d^t$ distinct rows appear the same number of times. In multiclass learning, a certain type of orthogonal arrays, known as Hadamard matrix, is found to be useful [48]. This binary square matrix has the following properties: $m = \sigma$ and $d = t = 2$. Table 2 presents an example of such a matrix. Note that any possible combination of any two columns of the matrix appears exactly the same number of times. Except for the first row, all the other rows have a constant Hamming weight of 4. In fact, it has been shown that Hadamard matrices are equivalent to certain constant weight codes [51]. In particular, a Hadamard matrix of order $m$ is equivalent to a constant weight code with $m$ columns and a constant Hamming weight of $k = m/2$. However, there is no way to set the Hamming weight to any desired value (such as $m = 11, k = 3$). Moreover, it should also be mentioned that constructing a Hadamard matrix is not a simple task and in some cases is even impossible.

On the other hand, the literature focuses on $(m, d, k)$ constant weight codes with $m$ columns, a constant weight of $k$ and a minimum Hamming distance of $d$ between any two rows. These matrices are not always suitable for our goal as they do not guarantee that all pairs of labels are covered. For example, Table 3 illustrates the matrix $A(19,10,8)$. One can note that the labels $\lambda_1$ and $\lambda_2$ are not covered (i.e. there is no row in which both $\lambda_1$ and $\lambda_2$ are equal to 1). Nevertheless, we assume that it is possible to construct qualified matrices by revising existing algorithms that were designed to construct constant weight codes (Östergård [31] provides an updated classification of such algorithms). Alternatively, one can use readymade constant weight codes as a starting point for searching the qualified matrices. These ideas are beyond the scope of this article, and we leave them for future research.

**Table 2**: Illustration of a Hadamard matrix ($m = \sigma$ and $d = t = 2$).

| $\lambda_1$ | $\lambda_2$ | $\lambda_3$ | $\lambda_4$ | $\lambda_5$ | $\lambda_6$ | $\lambda_7$ | $\lambda_8$ |
|---|---|---|---|---|---|---|---|
| 0 | 0 | 0 | 0 | 0 | 0 | 0 | 0 |
| 1 | 1 | 1 | 0 | 1 | 0 | 0 | 0 |
| 1 | 0 | 1 | 1 | 0 | 1 | 0 | 0 |
| 1 | 0 | 0 | 1 | 1 | 0 | 1 | 0 |
| 1 | 0 | 0 | 0 | 1 | 1 | 0 | 1 |
| 1 | 1 | 0 | 0 | 0 | 1 | 1 | 0 |
| 1 | 0 | 1 | 0 | 0 | 0 | 1 | 1 |
| 1 | 1 | 0 | 1 | 0 | 0 | 0 | 1 |

**Table 3**: Illustration of constant weight code $A(19,10,8)$

| λ1 | λ2 | λ3 | λ4 | λ5 | λ6 | λ7 | λ8 | λ9 | λ10 | λ11 | λ12 | λ13 | λ14 | λ15 | λ16 | λ17 | λ18 | λ19 |
|---|---|---|---|---|---|---|---|---|---|---|---|---|---|---|---|---|---|---|
| 1 | 0 | 0 | 0 | 1 | 1 | 0 | 1 | 0 | 0 | 0 | 1 | 1 | 0 | 0 | 0 | 1 | 0 | 1 |
| 1 | 0 | 0 | 0 | 1 | 1 | 0 | 0 | 1 | 0 | 0 | 0 | 0 | 1 | 1 | 1 | 0 | 1 | 0 |
| 1 | 0 | 0 | 1 | 0 | 0 | 1 | 0 | 0 | 1 | 0 | 0 | 0 | 1 | 1 | 0 | 1 | 0 | 1 |
| 1 | 0 | 0 | 1 | 0 | 0 | 1 | 0 | 0 | 0 | 1 | 1 | 1 | 0 | 0 | 1 | 0 | 1 | 0 |
| 0 | 1 | 0 | 1 | 0 | 1 | 0 | 0 | 1 | 1 | 0 | 1 | 0 | 0 | 0 | 1 | 1 | 0 | 0 |
| 0 | 1 | 0 | 0 | 1 | 0 | 1 | 0 | 1 | 1 | 0 | 0 | 1 | 0 | 0 | 0 | 0 | 1 | 1 |
| 0 | 1 | 0 | 1 | 0 | 1 | 0 | 1 | 0 | 0 | 1 | 0 | 0 | 1 | 0 | 0 | 0 | 1 | 1 |
| 0 | 1 | 0 | 0 | 1 | 0 | 1 | 1 | 0 | 0 | 1 | 0 | 0 | 0 | 1 | 1 | 1 | 0 | 0 |
| 0 | 0 | 1 | 0 | 0 | 1 | 1 | 1 | 0 | 1 | 0 | 0 | 1 | 1 | 0 | 1 | 0 | 0 | 0 |
| 0 | 0 | 1 | 1 | 1 | 0 | 0 | 0 | 1 | 0 | 1 | 0 | 1 | 1 | 0 | 0 | 1 | 0 | 0 |
| 0 | 0 | 1 | 1 | 1 | 0 | 0 | 1 | 0 | 1 | 0 | 1 | 0 | 0 | 1 | 0 | 0 | 1 | 0 |

In RAKEL the classification of a new sample $x$ is performed by the steps that are listed in Algorithm 2. Namely, each ensemble member returns a binary decision about the relevance of each label in $L$. The decisions are averaged for each label, and the labels whose average exceeds a given threshold $t$ are associated with $x$.

The complexities of both the construction and classification are linear with respect to the number of ensemble members (rows in the matrix $M$), as in most ensemble methods, and the total complexity depends on the complexity of the inducer that is used. The number of members can either be given as a parameter or it can be determined according to the given subset selection criteria. In the latter case, this number can be determined or estimated using combinatorial analysis. We provide such an estimate for the INLAC algorithm in Section 4.1.

---

**Algorithm 2: RAKEL classification phase**

*Input*: new instance– $x$,
        An ensemble of LP classifiers– $h_i$,
        $k$-labelset matrix – $M$

*Output*: Multi-label classification vector

1. **for** $j \leftarrow 1$ to $|L|$ **do**
2.     $Sum_j \leftarrow 0$;
3.     $Votes_j \leftarrow 0$;
4. **end for**
5. **for** $i \leftarrow 1$ to the number of rows in $M$ **do**
6.     $Y_i \leftarrow$ a labelset selected from $M$;
7.     **for** each labels $\lambda_j \in Y_i$ **do**
8.         $Sum_j \leftarrow Sum_j + h_i(x, \lambda_j)$;
9.         $Votes_j \leftarrow Votes_j + 1$;
10.     **end for**
11. **end for**
12. **for** $j \leftarrow 1$ to $|L|$ **do**
13.     $Avg_j \leftarrow Sum_j / Votes_j$
14.     **if** $Avg_j > t$ **then**
15.         $Result_j \leftarrow 1$;
16.     **else**
17.         $Result_j \leftarrow 0$;
18.     **end if**
19. **end for**

## 2.3 RAKEL++

In addition to the original implementation of RAKEL, in this paper we also examine a variant of RAKEL, denoted as RAKEL++ which implements two additional features:

1. *Incorporation of the classification confidence* - It has already been shown that for multi-class problems, error-correcting output codes perform much better when confidence is taken into consideration [1]. The original RAKEL implementation uses a simple voting scheme and ignores the confidence values attached to the predictions provided by the base-classifiers. In RAKEL++, instead of voting, the confidence values are taken into account by thresholding the average of the probabilities provided by the base-classifier for each label.

2. *Data-driven derivation of the threshold $t$* - Instead of manually setting up the threshold value by the user, RAKEL++ determines $t$ according to the dataset in hand by employing five-fold cross validation (see [15] for details). The value of $t$ is set to the average of the threshold values obtained by this cross validation procedure.

All the methods presented in the following sections use RAKEL++. From the practitioner's point of view, revising RAKEL to support the new features of RAKEL++ is very easy in the MULAN[1] package version 1.3. The confidence values are collected by the RAKEL implementation in the package when the base-classifier provides this information. In other cases the classifiers can be retrofitted to produce confidence values [30]. Moreover, the *OneThreshold* class can be used for finding the best threshold value. It operates in two stages. First, it evaluates the target measure using cross-validation at equally spaced points based on the user-defined step size (such as 0.1). Then when the proximity of the optimal threshold value is found, it uses fine-grained steps (such as 0.01) to converge to a smaller optimum interval. Following Tsoumakas and Vlahavas [45], the cross validation procedure first tests 9 different threshold values ranging from 0.1 to 0.9 in 0.1 steps. This procedure can be improved by using more efficient line search methods such as the golden section search method [27]. Moreover, in order to improve the efficiency of the evaluation procedure one can follow the procedure presented by Kohavi and John [28]. Instead of using a fixed number of cross validation folds, they suggest to repeat the cross validation procedure several times. The number of repetitions is determined on the fly by looking at the standard deviation of the estimated measure. If the standard deviation of the estimated measure is above certain value (for example 1%) another cross-validation is performed. Furthermore, in order to prevent the internal cross-validation from being too intensive, the CV procedure is performed only on a portion (10%) of the RAKEL's ensemble size when the ensemble size is indeed large enough (i.e. $\sigma > 100$).

## 2.4 Other Ensemble Algorithms

In this section we briefly review several recently proposed ensemble algorithms for multi-label classification. The data sparseness problem of the LP approach was addressed in [34]. The authors propose Pruned Sets (PS) and Ensemble of Pruned Sets (EPS) methods in order to focus on the most important correlations. This is achieved by pruning away examples with infrequently occurring label sets. Some of the pruned examples are then partially reintroduced into the data by decomposing them into more frequently occurring label subsets. Fi-nally, a process similar to the regular LP approach is applied to the new dataset. The authors show empirically that the proposed methods are often superior to other multi-label methods. However, these methods are likely to be inefficient in domains which contain a large percentage of distinct label combinations where examples are evenly distributed over those combinations [34]. Another limitation of the PS and EPS methods is the need to balance the trade-off between information loss (caused by pruning training examples) and adding too many decomposed examples with smaller label sets. For this purpose there is a need to choose some non-trivial parameter values before applying the algorithm or, alternatively, to perform calibration tests for parameters adjustment. Another limitation is that the inter-label correlations within the decomposed label sets are not considered.

---

[1] http://mulan.sourceforge.net/

Another approach for multi-label classification in domains that contain a large number of labels was proposed by Tsoumakas et al. [42]. The proposed algorithm (HOMER) organizes all labels in a tree-shaped hierarchy where each node contains a set of labels that is substantially smaller than the entire set of labels. A multi-label classifier is then constructed at each non-leaf node, following the BR approach. The multi-label classification is performed recursively, starting from the root and proceeding to the child nodes only if the child's labels are among those predicted by the parent's classifier. One of the main HOMER processes is the clustering of the label set into disjoint subsets so that similar labels are placed together. This is accomplished by applying a balanced $k$-means clustering algorithm to the label part of the data. However, this approach also ignores possible correlations among the labels within each tree node.

A recent paper argues in defense of the BR method [35]. It presents a method for chaining binary classifiers – Classifier Chains (CC) – in a way that overcomes the label independence assumption of BR. According to the proposed method, a single binary classifier is associated with each one of the predefined labels in the dataset and all these classifiers are linked in an ordered chain. The feature space of each classifier in the chain is extended with the 0/1 label associations of all previous classifiers. Thus, each classification decision for a certain label in the chain is augmented by all prior binary relevance predictions in the chain. In this manner correlations among labels are considered. The CC method has been shown to improve the classification accuracy of the BR method on a number of regular (not large-size) datasets. One of the disadvantages of this method, noted by authors, is that the order of the chain itself has an effect on accuracy. This can be solved either by a heuristic for selecting the order of the chain members or by using an ensemble of chain classifiers. Any of these solutions increases the required computation time. Another disadvantage of this approach is that in datasets that contain many features the effect of the proposed label augmentation to the large feature space is very small. It might even be neglected in datasets where the number of features is much higher than the number of labels.

Recently, a probabilistic extension of the CC algorithm was proposed [11]. According to the probabilistic classifier chains (PCC) approach, the conditional probability of each label combination is computed using the product rule of probability. In order to estimate the joint distribution of labels, a model is constructed for each label based on a feature space augmented by previous labels as additional attributes. The classification prediction is then explicitly derived from the calculated join distributions.

A few works on multi-label learning have directly identified dependent labels explicitly from the dataset. One such method where the degree of label correlation is explicitly measured was presented recently in [44]. In this paper, the authors use stacking of BR classifiers to alleviate the label correlations problem [46]. The idea in stacking is to train a second (or meta) level of models that consider as input the output of all first (or base) level models. In this way, correlations between labels are modeled by a meta-level classifier [52]. To avoid the noise that may be introduced by modeling uncorrelated labels in the meta-level, the authors prune models participating in the stacking process by explicitly measuring the degree of label correlation using the *phi* coefficient. They empirically showed that detected correlations are meaningful and useful. The main disadvantage of this method is that the identified inter-label correlations are utilized only by the meta-level classifier.

A recent paper by Zhang and Zhang [49] exploited conditional dependencies among labels. For this purpose the authors construct a Bayesian network to represent the joint probability of all labels conditioned by the feature space such that dependency relations among labels are explicitly expressed by the network structure. Zhang and Zhang [49] learn the network structure from classification errors of independent binary models which are constructed for all labels. Next, a new binary classifier is constructed for each label by augmenting its parent labels in the network to the feature space (similarly to the Classifier Chains approach). The labels of new examples are predicted using these classifiers where the ordering of the labels is implied by the Bayesian network structure. Zhang and Zhang [49] showed empirically that their method is highly comparable to the state-of-the-art approaches over a range of datasets using various multi-label evaluation measures. Note that the augmented parent labels have an equal influence as the rest of the features during the construction of the new classifiers. This may considerably reduce the potential benefit from utilizing the discovered inter-label dependencies.

Tenenboim et al. [39] proposed to discover existing dependencies among labels prior to the construction of any classifier and then to use the discovered dependencies to construct a multi-label classifier. Specifically, they defined methods for estimating the conditional and unconditional dependencies between labels in a given training set and then apply a new algorithm that combines the LP and BR methods to the results of each one of the dependence identification methods. The empirical results show that in many cases this approach outperforms many existing multi-label classification methods.

## 3 The Set Covering Problem framework for Labelset Selection

The set cover problem (SCP) is one of the 21 problems that were originally shown to be NP-complete by Karp [26]. In its simplest form, the problem is described as follows: given a set $\Omega$, a family $\Sigma$ of subsets of $\Omega$ and an integer $p$, does $\Sigma$ contain $p$ subsets which cover $\Omega$ i.e. whose union is equal to $\Omega$. When each element is covered exactly once, the cover is called a *prime cover*. The optimization version of the SCP problem appears in many areas such as scheduling and is known to be NP-hard [17, 26]. The general optimization SCP assigns a weight to each subset and looks for a cover whose total weight is minimal. This version is called the *weighted set cover problem* (WSCP). A particular case of the WSCP limits the size of each subset by a given value $k$ and is referred to as the *weighted k-set cover problem*. This formulation is of particular interest to the problem of $k$-labelset selection due to the need to limit the size of each labelset by a moderate value $k$ (Section 2).

### 3.1 WSCP as a Zero-One Integer Programming (IP) Problem

In the following we give a formal description of the weighted set cover problem as a zero-one integer programming problem. Let $\Omega = \{e_i\}_{i=1}^{m}$ be a set containing $m$ elements and let $\Sigma = \{S_i\}_{i=1}^{n}$ be a group of $n$ subsets of $\Omega$, where $|S_i| \geq 1$ and $w_i$ is a weight associated with $S_i$. We say that $I \subseteq \{1, \dots, n\}$ is a cover of $\Omega$ if $\Omega = \bigcup_{j \in I} S_j$.

The cost of the cover $I$ is given by $Z(I) = \sum_{j \in I} C(S_j)$ where $C(S_i)$ measures the cost of including $S_i$ in the cover $I$. Usually, $C(S_i)$ is a function of $w_i$.

The weighted set cover problem can be formulated as a *zero-one integer programming* (IP) problem [18] as follows:

$$\text{minimize } Z(I) = \sum_{j=1}^{n} C(S_j) \cdot x_j$$

$$\text{subject to } \sum_{j=1}^{n} a_{ij} \cdot x_j \geq 1, \quad i = 1, \dots, m$$

$$x_j \in \{0,1\}, \quad j = 1, \dots, n$$

where

$$x_j = \begin{cases} 1 & \text{if } j \in I \ (S_j \text{ is in the cover of } \Omega) \\ 0 & \text{otherwise} \end{cases}$$

and

$$a_{ij} = \begin{cases} 1 & \text{if } e_i \in S_j \\ 0 & \text{otherwise} \end{cases}.$$

This zero-one IP problem can be directly solved by enumerating all $2^n$ zero-one vectors of $I$ [29]. The algorithm performs the enumeration by iterating through the following steps:

1. Choose a free variable $x_j$ and fix it to the value 1.
2. Enumerate each of the completions of the partial solution.
3. Fix the variable $x_j$ to the value 0.
4. Repeat the process for the sub-problem with $x_j$ equal to 0.

The algorithm uses an effective branching test to choose which free variable $x_j$ to set to 1 in such a way that the sum of the absolute values of the amount by which all constraints are violated is maximally reduced. Since the algorithm enumerates implicitly all the possible vectors of $I$, it guarantees an optimal solution if one exists. Nevertheless, this algorithm is only practical for small scale problems due its exponential time complexity. Consequently, a heuristic algorithm is required in order to find a sufficiently good solution for large scale problems. Using the AMPL (A Mathematical Programming Language) environment together with the CPLEX package, we were able to find an exact solution only for small scale SCP problems which correspond to multilabel problems of less than 15 labels.

## 3.2 Approximation algorithms for solving the WSCP

Many heuristics have been proposed for approximating the solution of the SCP. Grossman and Wool [21] conducted a comparative study of nine approximation algorithms for the *unweighted set covering problem*. The authors showed that the empirical performances of the *randomized greedy algorithm* [7] and the *randomized rounding algorithm* [33] are better than the performance of the other algorithms. In each iteration of the randomized greedy algorithm, the variable that appears in the largest number of unsatisfied inequalities is picked (ties are broken at random). The randomized rounding algorithm first solves the fractional version of the problem, and then uses randomization to obtain an approximate solution for the integer problem. Namely, each frac-tional value is multiplied by a scaling factor greater than 1. Then a biased coin is tossed for each variable and if the coin is "1" the variable is picked.

In a recent paper [20] a comparison is conducted between the theoretical and experimental performance of several approximation algorithms. The results showed that the greedy algorithm performed extremely well on all of the tested instances when it was applied to the unweighted and weighted versions of the SCP. Moreover, the results indicated that changing between the weighted and unweighted settings has very little effect on the greedy algorithm compared to the rounding algorithm. Given the findings of this study, we chose to focus on the *greedy* algorithm.

The basic idea of the *greedy* algorithm [7] is to pick the subset from Σ that covers the highest number of uncovered elements at the lowest cost. The process is repeated until all of the elements in Ω are covered. This algorithm is described in Algorithm 3. The algorithm initializes the solution $I$ to an empty set. Then the algorithm finds the subset $S_q$ whose cost is minimal, where the cost of each subset $S_j$ is calculated as the ratio between its associated weight and its size. Next, the subset $S_q$ is added to the solution $I$, and the elements covered by $S_q$ are removed from all of the subsets in Σ. This process is repeated until the subsets in $I$ cover all of the elements in Ω. Note that due to the element removal step, the size of each subset $S_j$ is updated to be the number of unique, *uncovered* elements contained in it.

---

**Algorithm 3: The Greedy Algorithm for weighted set cover**

**Input**: Finite set $\Omega = \{e_i\}_{i=1}^{m}$
A group of subsets of Ω ; $\Sigma = \{S_i\}_{i=1}^{n}$
Subset weights $W = \{w_i\}_{i=1}^{n}$
A cost function $C: \Sigma \rightarrow \mathbb{R}$
**Output**: A set covering $I$

1. $I \leftarrow \phi$
2. **for** $i$=1 **to** $n$ **do**
3.    $S_i^* \leftarrow S_i$
4. **end for**
5. **while** $\Omega \neq \bigcup_{i \in I} S_i$ **do**
6.    $q \leftarrow \arg\min_{j: S_j^* \neq \phi} \frac{w_j}{|S_j^*|}$
7.    $I \leftarrow I \cup \{q\}$
8.    **for** $l$=1 **to** $n$ **do**
9.       $S_l^* \leftarrow S_l^* \backslash S_q$
10.    **end for**
11. **end while**

---

## 4 The Labelset Selection Problem as a Set Covering Problem

We propose three strategies for selecting *k*-labelsets in order to construct a multi-label ensemble classifier. Each proposed strategy is formulated as a *k*-set cover problem having its own cost function and Ω, Σ sets. Each proposed strategy employs additional heuristic criteria to improve the predictive performance of the produced ensemble classifier.

Contrary to the RAKEL algorithm in which the ensemble size $\sigma$ is determined by the user, in our case the ensemble size can be either determined by the user or it can be determined according to the outcome of the SCP solution. Following the formal SCP description in Section 3.1, the ensemble size is simply $\sigma = \sum_{i=1}^{n} x_i$. We are looking for the smallest $\sigma$ that satisfies all criteria.

## 4.1 The INter-LAbel Correlations (INLAC) Strategy

In many datasets, some of the labels are correlated. In the task of classifying movies in genres, a movie that is labeled with the genre "Animation" has a high chance to be also labeled with the genre "Adventure" but relatively low chance to be additionally labeled with the genre "Horror". It is important that the codeword matrix will cover the positively correlated pair ("Animation", "Adventure") as well as the negatively correlated pair ("Animation","Horror"). In Section 4.4 we use the notion of statistical dependency to examine which labels it is important to cover.

The need to cover inter-label correlations motivates a strategy for constructing a cover matrix that includes all possible pairs and triplets. Accordingly, the algorithm needs to receive as input $r$ - the *order* of inter-label correlations that are sought after. Setting $r = 2$ covers inter-label correlations between pairs of labels. In order to cover inter-label correlations among triplets, $r$ is set to 3 and so forth.

The SCP formulation of this strategy initializes the sets $\Omega$ and $\Sigma$ with all $r$-labelsets and $k$-labelsets of $L$, respectively. A certain $k$-labelset $S_j$ covers a certain $r$-labelset $r_i$ if $r_i \subseteq S_j$. For example the triplets $r_1 = \{\lambda_4, \lambda_8, \lambda_9\}$ and $r_2 = \{\lambda_4, \lambda_6, \lambda_8\}$ are all covered by $S_j = \{\lambda_4, \lambda_6, \lambda_8, \lambda_9\}$. Thus, we construct a cover of all $r$-labelsets instead of individual labels. This will also cover all individual labels. Note that each label is always covered together with $r - 1$ other labels. This way, if the $r$ labels that are covered in a given iteration are correlated, they will be included in the cover. We set a uniform weight to each of the $k$-labelsets, since there is no a-priori preference of one $k$-labelset over another. The cost function $C(S_i)$ is defined as the negation of the number of uncovered $r$-labelsets that are covered by $S_i$ (the negation is taken in order to be consistent with the formulation of the SCP as a minimization problem). Note that the lowest possible cost is $-\binom{k}{r}$. The steps for obtaining a solution employing the INLAC strategy are given in Algorithm 4.

**Algorithm 4: The INLAC Greedy Algorithm**

*Input*: The set of $r$-labelsets of $L$ ; $\Omega = \{r_i\}_{i=1}^{o}$
   The group of $k$-labelsets of $L$ ; $\Sigma = \{S_i\}_{i=1}^{n}$
   The ensemble size $\sigma$
*Output*: A set covering $I$

1. $I \leftarrow \phi$
2. **for** $i$=1 **to** $n$ **do**
3.    $S_i^* \leftarrow$ all $r$-labelsets that are covered by $S_i$
4. **end for**

5. **Do**
6.    $q \leftarrow \arg\max_j(|S_j^*|)$
7.    $I \leftarrow I \cup \{S_q\}$
8.    $\Omega = \Omega - S_q^*$
9.    **for** $l$=1 **to** $n$ **do**
10.       $S_l^* \leftarrow S_l^* \backslash S_q$
11.    **end for**
12. **Until** ($\sigma = \infty$ AND $\Omega = \emptyset$ ) OR ($|I| = \sigma$)

Initially, we set $S_i^*$ to all $r$-labelsets that are covered by the $k$-labelset $S_i$. Then, we iteratively find the $k$-labelset $S_q$ that covers the highest number of uncovered $r$-labelsets. The $r$-labelsets that are covered by $S_q$ are then removed from the set of all $r$-labelsets $\{S_i^*\}$. We use the binary matrix representation that was described in Section 2 in order to illustrate a covering. Table 4 displays the result of this algorithm for a set of $m = 4$ labels which is covered by 3-labelsets ($k = 3$) including all pair-wise correlations ($r = 2$).

**Table 4:** Matrix set of $m = |L| = 4$ labels which is covered by 3-labelsets ($k = 3$) including all pair-wise correlations.

|       | $\lambda_1$ | $\lambda_2$ | $\lambda_3$ | $\lambda_4$ |
|-------|---|---|---|---|
| $S_1$ | 1 | 1 | 1 | 0 |
| $S_2$ | 1 | 1 | 0 | 1 |
| $S_3$ | 1 | 0 | 1 | 1 |

Inter-label correlation can also be covered by RAKEL. However, since the $k$-labelsets are randomly chosen, some correlations may be missed. Lemma 1 quantifies the ability of randomly chosen $k$-labelsets to cover inter correlations between pairs of labels.

**Lemma 1**: The probability that all pairs in a label set $L = \{\lambda_i\}_{i=1}^{m}$ will be covered by $\sigma$ random $k$-labelsets is bounded by:

$$p \leq 1 - \frac{2}{h+1} S_1 + \frac{2}{h(h+1)} S_2$$

where:

$$S_1 = \binom{m}{2} \left( \frac{\binom{m-2}{k} + 2\binom{m-2}{k-1}}{\binom{m}{k}} \right)^{\sigma} = \frac{m(m-1)}{2} \cdot \left( \frac{(m-k)(m+k-1)}{m(m-1)} \right)^{\sigma}$$

$$S_2 = 3\binom{m}{4} \left( \frac{\binom{m-4}{k} + 4\binom{m-4}{k-1} + 4\binom{m-4}{k-2}}{\binom{m}{k}} \right)^{\sigma} + 3\binom{m}{3} \left( \frac{\binom{m-3}{k} + 3\binom{m-3}{k-1} + \binom{m-3}{k-2}}{\binom{m}{k}} \right)^{\sigma}$$

and

$$h = 1 + \left\lfloor \frac{2 S_2}{S_1} \right\rfloor.$$

The proof of *Lemma 1* is provided in the appendix. For the sake of simplicity, *Lemma 1* assumes random selection of $k$-labelsets with replacement while the RAKEL algorithm selects labelsets without replacement. Figure 1 presents the coverage probability $p$ ($y$-axis), with respect to the number of ensemble members $\sigma$ ($x$-axis), number of labels ($L$) and subset size ($k$). As expected, the coverage probability asymptotically converges to 1 as $\sigma$ increases. It can be seen that: (a) The convergence rate is slower for larger values of $|L|$; and (b) The convergence is faster for higher values of $k$. Nevertheless, increase in $k$ also increases the complexity of the classifier. From a computational cost perspective, smaller ensembles are preferred. Moreover, larger ensembles do not necessarily improve predictive performance (see for instance [50]). Yet there are other reasons for preferring small ensembles:

- Smaller ensembles require less memory for storing their members.
- Smaller ensembles are considered to be more comprehensible to the end-users.
- Compact ensembles provide higher classification speed. It is particularly crucial in real-time applications, such as worm detection in PC's, that not only need to pursue the highest possible accuracy, but are also required to respond as fast as possible. The focus of a recent paper which uses the ECOC approach for solving multi-class problems is minimizing the number of classifiers that are used during the classification [32].

Algorithm 4 is designed to work in two different modes. In the first mode, the algorithm constructs an ensemble whose size is set explicitly by the user via the parameter σ. In the second mode (σ = ∞) the algorithm stops when all *r*-labelsets are covered. If σ is set to a value that is greater than the size needed to cover all *r*-labelsets, then the surplus elements in *I* are arbitrarily selected.

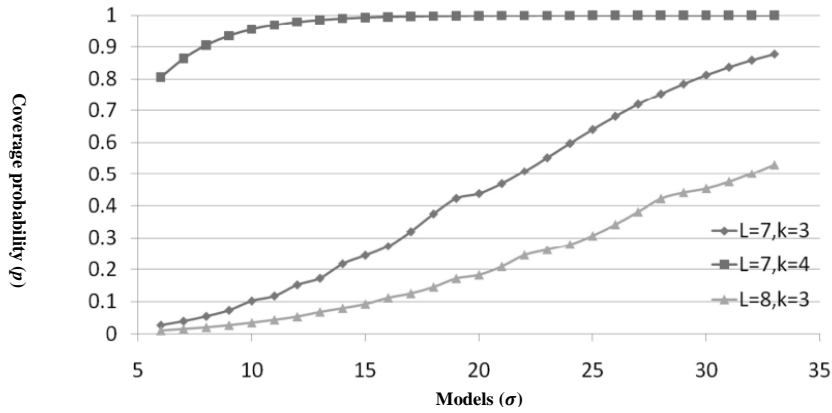

**Figure 1:** Coverage probability *p* for covering $m = |L|$ labels by *k*-labelsets using σ ensemble members.

**Complexity Analysis**

The dominant steps in the algorithm are 3, 6 and 10. Step 3 requires $O\left(\binom{m}{k} \cdot \binom{k}{r}\right) = O(m^k \cdot k^r)$ time and space. Step 6 requires $O\left(\binom{m}{k} \cdot \binom{m}{r}\right) = O(m^{k+r})$ and the entire loop takes $O(\sigma \cdot m^{k+r})$. Step 10 requires $O\left(\binom{m}{k}\right) = O(m^k)$. Since $k^r \ll m^k$ and $\sigma \ll m^k$, the overall complexity is $O(m^{k+r})$.

The storage complexity is $O(m^k + m^r)$, since we need to maintain all *k*-labelsets and *r*-labelsets in memory. It is important to emphasize that the labelsets preparation can be performed offline and is done only once for each given values of $m, k, r$ and σ.

**Analysis of the cover size**

Lemma 2 estimates how many *k*-labelsets are required to cover all inter-label correlations of order *r*.

**Lemma 2**

Given a set of labels $L = \{\lambda_i\}_{i=1}^m$, the number of *k*-labelset required to cover all inter-label correlations of der *r* is given by:

$$\frac{\binom{m}{r}}{\binom{k}{r}} \leq \sigma \leq \frac{m}{k} + \binom{m}{r} - \frac{m}{k}\binom{k}{r}$$

**Proof of Lemma 2**

We represent every inter-label correlation by an *r*-labelset. Each *k*-labelset can cover at most $\binom{k}{r}$ new previously uncovered *r*-labelsets. For complete coverage we need to cover $\binom{m}{r}$ *r*-labelsets. Therefore, the size of $\sigma$ must be at least $\frac{\binom{m}{r}}{\binom{k}{r}}$.

On the other hand, we can construct at least $\frac{m}{k}$ mutually exclusive *k*-labelsets such that each one of them covers $\binom{k}{r}$ new *r*-labelsets (for the sake of simplicity, we assume that $m \bmod k = 0$). The rest of the *k*-labelsets will cover at least one *r*-labelset each (otherwise this labelset will not be included in the cover). There are $\binom{m}{r} - \frac{m}{k}\binom{k}{r}$ *r*- labelsets that will be covered in this manner. Therefore $\sigma$ cannot exceed:

$$\frac{m}{k} + \binom{m}{r} - \frac{m}{k}\binom{k}{r} \blacksquare$$

It should be emphasized that the matrix representing the constructed cover is different from the covering arrays and orthogonal arrays [25]. It is different since each line in the matrix is required to represent a *k*-labelset as in RAKEL. This ensures that the training complexity of all classifiers will be similar. If the value of *k* is not identical for all the labelsets (=classifiers), then one classifier can be much more difficult to train than the others (see for example [22] for an explanation how the number of classes affects the training).

### 4.2 The BAlanced Label COntribution (BALCO) Strategy

Recall that the classification algorithm (Algorithm 2) employs a voting scheme in order to derive the prediction of a new sample. In a randomly constructed matrix (as in RAKEL) certain labels may appear more than others. Assuming that all labels are equally difficult to be learnt, there is a greater chance that we misclassify instances of undersampled labels because these labels are described by a smaller number of classifiers. To take it to the extreme if a certain label does not appear at all in the matrix, then we cannot decide if a certain instance belongs to that label. Let *P* denote the probability that at least one label out of *m* labels is not represented by none of the $\sigma$ randomly selected *k*-labelsets. Employing the same proof strategy that we used in Lemma 1, it can be shown that

$$P \geq \frac{2}{h+1}S_1 - \frac{2}{h(h+1)}S_2$$

where:

$$h = 1 + \left\lfloor \frac{2S_2}{S_1} \right\rfloor; S_1 = \frac{(m-k)^\sigma}{m^{\sigma-1}}; S_2 = \frac{((m-k)(m-k-1))^\sigma}{2(m(m-1))^{\sigma-1}}$$

For small values of $\sigma$, this probability cannot be neglected. For example, given an ensemble of $\sigma = 100$ labelsets, $k = 3$ and number of labels $m = 100$, the misrepresentation probability is greater than 0.86.

In order to prevent any misrepresentation, one must select the labelsets such that all labels will be evenly represented. This labelset selection problem can be formulated as a set cover problem, such that the set $\Omega$ will be composed of the set of all labels $L$ while the set $\Sigma$ will be initialized to all $k$-labelsets of $L$. The subsets weights will all be set to 1. We define the frequency of a label as the number of labelsets in the cover that include it. Let $imb(I)$ be the imbalance level of a cover $I$, which we define as the frequency difference between the most frequent label and the least frequent label in $I$. Accordingly, the cost function $C(S_i)$ is defined as $imb(S_i \cup I)$, which penalizes a labelset $S_i$ according to the amount of imbalance it adds to the cover. Algorithm 5 lists the steps for obtaining a cover according to the BALCO strategy.

---

**Algorithm 5: The BALCO Greedy Algorithm**

**Input**: The set of labels $\Omega = L = \{\lambda_i\}_{i=1}^m$
  The group of $k$-labelsets of $\Omega$ ; $\Sigma = \{S_i\}_{i=1}^n$
  A cost function $C: \Sigma \to \mathbb{R}$
  The size of the cover $\sigma$
**Output**: A set covering $I$

1. $I \leftarrow \phi$
2. **while** $|I| < \sigma$ **do**
3.   $q \leftarrow \arg\min_i C(S_i)$
4.   $I \leftarrow I \cup \{q\}$
5.   $\Sigma = \Sigma - S_q$
6. **end while**

---

**Complexity Analysis**

The dominant step in the analysis is step 3, since it searches all $\binom{m}{k}$ possible $k$-labelsets. Consequently, the time and space complexities of the algorithm are $O(\sigma m^k)$.

**Analysis of the cover size**

Recall that the complexity of the ensemble is linear in the number of ensemble members. It is easy to see that at any given stage $imb(I) \leq 1$ and $imb(I) = 0$ when $gcd(|I| \cdot k, n) = n$ for $|I| \geq 1$ where $gcd$ is the greatest common divisor. A prime cover, i.e. a cover where each label is covered exactly once, can only be obtained if $(m \bmod k) = 0$. If $(m \bmod k) \neq 0$, we need $\sigma = lcm(k, m)/k$ $k$-labelset to obtain a balanced cover where $lcm$ is the least common multiple.

## 4.3 The BALANced label contribution inter-label CORrelations (BALANCOR) strategy

The BALCO and INLAC strategies enforce criteria that are *both* important for obtaining an ensemble classifier whose predictive performance is better than RAKEL. Accordingly, a hybrid strategy that simultaneously enforces both criteria should benefit from their advantages. When using the SCP as the algorithmic framework, this strategy can be enforced by merely changing the cost function in step 6 of Algorithm 4 to $q \leftarrow \arg\max_j \left(|S_j^*| - imb(I \cup S_j)\right)$. Thus, the cost is composed of the number of $r$-labelsets that are covered and it is being penalized by the contribution to the imbalance of the cover. The time and space complexity analysis is similar to the one in the INLAC strategy since calculating the imbalance in step 6 can be done in constant time for each inspected $k$-labelset.

### 4.4 Data Driven BALANCOR Strategy (DD-BALANCOR)

So far the matrix design methods ignored the dataset in hand. By examining the data, it should be possible to estimate which $k$-labelsets contribute more to the ensemble. One possibility is to revise the lableset selection criterion (step 6) in Algorithm 4 to take the labels dependency into account. However, implementing this approach will require a different construction process for every classification task which will prevent us from separating the matrix construction from the ensemble training. Therefore, we adopt a different approach. Given a matrix that was constructed using the BALANCOR strategy as described in Section 4.3, we add a pre-processing step prior to the ensemble training that adjusts the matrix according to the dataset in hand. Specifically, the adjustments will permute the labels according to a given objective function.

Note that a matrix of power $r < k$, covers all $r$-labelsets but only a portion of the $(r + 1)$-labelsets. Thus, a possible adjustment of the matrix can be to find a label permutation (which is equivalent to permuting the matrix columns) in which the highest number of $(r + 1)$ dependent labels is covered. Specifically, we aim to maximize the total sum of *weighted dependency levels* of all $(r + 1)$ labelsets covered by the matrix. The dependency level weight of a certain $(r + 1)$ labelset is set to the number of times its labels co-occur in the training set (i.e. the number of instances that correspond to this labelset). Recall the matrix presented in Table 4. If the data in hand implies that the labels in the triplet $\{\lambda_2, \lambda_3, \lambda_4\}$ are strongly dependent but the labels $\{\lambda_1, \lambda_2, \lambda_3\}$ are independent then a better columns ordering is given by the permutation $\{\lambda_2, \lambda_3, \lambda_4, \lambda_1\}$.

To this end, we first calculate the statistical dependencies among all $(r + 1)$ labelsets based on the training set. We apply the chi-square test for independence for all $(r + 1)$ labelsets. For example in case that $r = 2$ we evaluate the dependency among all label triplets: $\lambda_i, \lambda_j, \lambda_k$. Calculating the dependency begins by creating a three- way contingency table that stores the labels co-occurrences counts in the dataset in hand as illustrated in Table 5 (note that for the sake of presentation we are using two tables of two dimensions).

**Table 5** General contingency table for three labels

|  | $\lambda_i$ | |
|---|---|---|
|  | $\lambda_j$ | $\neg\lambda_j$ |
| $\lambda_k$ | $n_{1,1,1}$ | $n_{1,2,1}$ |
| $\neg\lambda_k$ | $n_{1,1,2}$ | $n_{1,2,2}$ |

|  | $\neg\lambda_i$ | |
|---|---|---|
|  | $\lambda_j$ | $\neg\lambda_j$ |
| $\lambda_k$ | $n_{2,1,1}$ | $n_{2,2,1}$ |
| $\neg\lambda_k$ | $n_{2,1,2}$ | $n_{2,2,2}$ |

Given the contingency table, the $\chi^2$ score can be computed as follows:

$$\chi^2 = \sum_i \sum_j \sum_k \frac{(E_{ijk} - n_{ijk})^2}{E_{ijk}}$$

where $E_{ijk}$ is the expected cell count, defined as:

$$E_{ijk} = \frac{\left(\sum_{j'=1}^{2}\sum_{k'=1}^{2} n_{ij'k'}\right)\left(\sum_{i'=1}^{2}\sum_{k'=1}^{2} n_{i'jk'}\right)\left(\sum_{i'=1}^{2}\sum_{j'=1}^{2} n_{i'j'k}\right)}{\left(\sum_{i'=1}^{2}\sum_{j'=1}^{2}\sum_{k'=1}^{2} n_{i'j'k'}\right)^2}$$

A high $\chi^2$ score indicates that the three labels are strongly dependent. Practically, there is no need to calculate the inter-dependency of all possible $(r + 1)$ labelsets. We take into account a certain $(r + 1)$ labelset if its labels co-occur at least 5 times in the dataset. The threshold value of 5 was chosen since it is used in statistics as a common practice for guarantying the correctness of the $\chi^2$ statistics. Specifically, the standard $\chi^2$ test should be used only if the total number of observations is greater than 40 and the expected frequency in each cell is at least 5 [9].

In order to search for the best permutation we employ a simulated annealing algorithm. We begin with a random permutation and calculate its merit i.e. the total sum of weighted dependency levels of all $(r + 1)$ labelsets covered by the matrix under this permutation. In every iteration, we look for an improved permutation which is a neighbor of the current permutation. We define a neighborhood to be the set of all permutations that can be obtained by interchanging the position of any pair of labels. According to this definition of neighborhood, each permutation has $\frac{m(m-1)}{2}$ neighbors. We select a neighbor if its merit is higher than that of the current permutation. Otherwise, the new neighbor is accepted with probability of $e^{-\Delta \cdot T_j}$, where $T_j$ is a parameter known as the temperature and $\Delta$ is the merit difference between the current permutation and its candidate neighbor. The temperature is high in the initial iterations to allow non-improving solutions to be accepted and then it decreases until it is close to zero. We used the following simple and common exponential schedule to update $T$: $T_j = \gamma T_{j-1}$ where the parameter $0 \leq \gamma \leq 1$ controls the rate of the decay (in this paper we used $\gamma = 0.85$). In either case, the selected new neighbor is set as the current permutation, its neighbors are generated and the process is repeated. In order to bound the number of iterations, we observed the number of iterations during which a significant improvement in the permutation merit took place for various values of $m$ and ensemble sizes $\sigma$. Fitting a regression equation, we recommend setting the number of iterations to:

$$N = \max\{1368 ln(m) + 12 ln(\sigma) - 2179, 2000\}.$$

# 5 Experimental Study

## 5.1 Experimental Setup

The three proposed strategies were empirically compared with RAKEL. We developed a software package in Matlab in order to generate various *k*-labelset covers of *m* labels. For the classification tasks, we used WEKA [16] and MULAN[2] (a software package for multilabel classification and ranking that is based on the WEKA framework). We used WEKA's SMO and J48 as our baseline classifiers for single-label classification in all ensemble models. The parameter values of both base-classifiers were set to the defaults provided by WEKA.

The following datasets[3] were used for the evaluation: *Scene*, *Emotions*, *Yeast*, *Slashdot*, *OHSUMED*, *Genbase*, *Medical*, *Enron* [14], *Delicous* and *MediaMill*. Table 6 presents certain properties of these datasets. The label cardinality is the average number of labels per example while the label density is the label cardinality divided by |L|.

**Table 6:** Properties of the datasets used in the experiments.

|  | Instances | | Attributes | | Labels | | |
|---|---|---|---|---|---|---|---|
| Dataset | Train | Test | Nominal | Numeric | Labels | Label Cardinality | Label Density |
| *Scene* | 2407 | CV | 0 | 294 | 6 | 1.074 | 0.179 |
| *Emotions* | 593 | CV | 0 | 72 | 6 | 1.869 | 0.311 |
| *Yeast* | 2417 | CV | 0 | 103 | 14 | 4.237 | 0.303 |
| *Slashdot* | 3782 | CV | 1079 | 0 | 22 | 1.18 | 0.041 |
| *OHSUMED* | 13929 | CV | 1002 | 0 | 23 | 1.66 | 0.082 |
| *Genbase* | 662 | CV | 1186 | 0 | 27 | 1.252 | 0.046 |
| *Medical* | 978 | CV | 1449 | 0 | 45 | 1.245 | 0.028 |
| *Enron* | 1702 | CV | 1001 | 0 | 53 | 3.378 | 0.064 |
| *Delicous* | 12920 | 3185 | 500 | 0 | 983 | 19.020 | 0.019 |
| *MediaMill* | 30993 | 12914 | 0 | 120 | 101 | 4.376 | 0.043 |

## 5.2 Evaluation Measures

In this paper we consider the most commonly used multi-label evaluation measures from (Tsoumakas and Vlahavas, 2007), namely multi-label example-based classification accuracy, subset-accuracy, Hamming loss, and label-based micro-averaged F-measure. Their formal definition and analysis are presented below.

Let $D$ be a multi-label evaluation dataset, consisting of $|D|$ multi-label examples $(X_i, Y_i)$, $i = 1 \ldots |D|$, $Y_i \subseteq [L]$. Let $h$ be a multi-label classifier.

Hamming loss computes the percentage of labels whose relevance is predicted incorrectly. For the two label subsets $A, B \subseteq [L]$, their Hamming distance is

$$\ell_{Ham}(A, B) = \frac{1}{L} \sum_{i=1}^{L} 1_{\{i \in A\}} \neq 1_{\{i \in B\}}. \tag{1}$$

Over all dataset examples the Hamming loss is averaged as follows:

---

[2] http://mulan.sourceforge.net/
[3]The datasets can be obtained from http://mulan.sourceforge.net/datasets.html and from http://meka.sourceforge.net/#datasets

$$Hamming\ loss(h, D) = \frac{1}{|D|} \sum_{i=1}^{|D|} \frac{Y_i \Delta h(X_i)}{|L|}$$

where $\Delta$ stands for the symmetric difference between two sets.

Hamming loss is very sensitive to the label set size *L*. It measures the percentage of incorrectly predicted labels both positive and negative. Thus, in cases where the percentage of positive labels is low relative to *L*, the low values of the Hamming loss measure do not give an indication of high predictive performance. Thus, as the empirical evaluation results demonstrate below, the accuracy of the classification algorithm on two datasets with similar Hamming loss values may vary from about 30 to above 70 percent (as, for example, in the "bibtex" and "medical" datasets). However, the Hamming loss measure can be useful for certain applications where errors of all types (i.e., incorrect prediction of negative labels and missing positive labels) are equally important.

Subset accuracy computes the number of exact predictions, i.e., when the predicted set of labels exactly matches the true set of labels. This measure is the opposite of the zero-one loss, which for the two binary vectors $A, B \subseteq [L]$ is defined as follows:

$$\ell_{01}(A, B) = \mathbf{1}_{\{A \neq B\}}. \tag{2}$$

Over all dataset examples the Subset accuracy is averaged as follows:

$$Subset\ accuracy(h, D) = \frac{1}{|D|} \sum_{i=1}^{|D|} I(Y_i = h(X_i)).$$

It should be noted also that subset accuracy is a very strict measure since it requires the predicted set of labels to be an exact match of the true set of labels, and equally penalizes predictions that may be almost correct and/or totally wrong. However, it can be useful for certain applications where classification is only one step in a chain of processes and the exact performance of the classifier is highly important (Vilar, 2004)

Accuracy computes the percentage of correctly predicted labels among all predicted and true labels. Accuracy averaged over all dataset examples is defined as follows:

$$Accuracy(h, D) = \frac{1}{|D|} \sum_{i=1}^{|D|} \frac{Y_i \cap h(X_i)}{Y_i \cup h(X_i)}.$$

Accuracy seems to be a more balanced measure and better indicator of an actual algorithm's predictive performance for most standard classification problems than Hamming loss and subset accuracy. However, it should be noted that it also is relatively sensitive to dataset label cardinality (average number of labels per example). This means that for two classification problems (i.e., datasets) of the same complexity, accuracy values would be lower in the dataset with the higher label cardinality. The empirical evaluation experiment below supports this conclusion (consider, for example accuracy and F-measure values on "emotions", "scene" and "yeast" datasets).

The F-measure is the harmonic mean between precision ($\pi$) and recall ($\rho$) and is commonly used in information retrieval. Precision and recall are defined as follows:

$$\pi_\lambda = \frac{TP_\lambda}{TP_\lambda + FP_\lambda}, \quad \rho_\lambda = \frac{TP_\lambda}{TP_\lambda + FN_\lambda},$$

where $TP_\lambda$, $FP_\lambda$ and $FN_\lambda$ stands for the number of true positives, false positives and false negatives correspondingly after binary evaluation for a label $\lambda$.

The micro-averaged precision and recall are calculated by summing over all individual decisions:

$$\pi = \frac{\sum_{\lambda=1}^{L} TP_\lambda}{\sum_{\lambda=1}^{L}(TP_\lambda + FP_\lambda)}, \qquad \rho = \frac{\sum_{\lambda=1}^{L} TP_\lambda}{\sum_{\lambda=1}^{L}(TP_\lambda + FN_\lambda)},$$

where $L$ is the number of labels. The micro-averaged F-measure score of the entire classification problem is then computed as:

$$F(micro - averaged) = \frac{2\pi\rho}{\pi + \rho}$$

Note that micro-averaged F-measure gives equal weight to each document and is therefore considered as an average over all the document/label pairs. It tends to be dominated by the classifier's performance on common categories and is less influenced by the classifier's performance on rare categories.

Of the various measures that are discussed here, the micro-averaged F-measure seems to be the most balanced and the least dependent on dataset properties. Thus it could be the most useful indicator of classifier general predictive performance for various classification problems. However it is more difficult for human interpretation, as it combines two other measures (precision and recall).

Summarizing the above analysis of some of the most commonly used evaluation measures, we conclude that accuracy and micro-averaged F-measure are better suited for general evaluation of algorithm performance for most regular multi-label classification problems while the Hamming loss and subset accuracy measures may be more appropriate for some specific multi-label classification problems.

Actually the accuracy measure, where the number of true labels among the predicted ones is important, can be considered as a "golden mean" between Hamming loss where all labels are equally important and subset-accuracy where only the whole set of positive labels is important. Thus, in this research we aim at improving the accuracy measure.

Some other evaluation measures, such as one-error, coverage, ranking loss and average precision, which are specially designed for multi-label ranking do exist (Schapire & Singer, 2000). This category of measures, known as ranking-based, is often used in the literature (although not directly related to multi-label classification), and is nicely presented in (Tsoumakas et al., 2010) among other publications. These measures are tailored for evaluation of specific-purpose ranking problems and are of a less interest for our research.

### 5.3 Evaluation Procedure

In order to estimate the generalization performance, a 10-fold cross-validation (CV) procedure was used. For each 10-fold cross-validation, the training set was randomly partitioned into 10 disjoint instances subsets. Each subset was utilized once as a test set and nine times as part of a training set. All algorithms were applied to the same cross-validation folds. We used train/test splits for evaluation of the two largest datasets (Delicious and MediaMill) as indicated in Table 6 since cross validation is too computational intensive in these cases.

In order to determine which algorithm performs best over multiple datasets, we followed the procedure proposed in [12]. We first used the adjusted Friedman test in order to test the null hypothesis that all methods perform the same and then the Nemenyi test to examine whether the new algorithm performs significantly better than existing algorithms.

The input parameters of the algorithm are:

- ***k*** – Labels subset size. Our parameter *k* corresponds to the user-specified parameter *k* – the size of the labelsets as defined in the RAKEL algorithm.
- ***σ*** – Ensemble size – defined according to the output of the minimum SCP method. The *σ* parameter corresponds to the ensemble size parameter in RAKEL. However, in contrast to RAKEL, the number of members in our ensemble is constant for each *k* and *m*, since it is defined by the solution of the BALANCOR algorithm. The ensemble size of the two other strategies has been adjusted to that of BALANCOR.
- ***r*** – Coverage power. In this paper we examine $r = 2$ and $r = 3$, i.e. ensembles that cover all pairs and all triplets of labels, respectively.
- ***t*** – Threshold – In RAKEL the value is set by the user. Recall that in RAKEL++ and in our derived methods, we use an internal cross validation procedure for selecting the most promising threshold value.

We compared the results achieved by the proposed strategies with those achieved by RAKEL where all algorithms used the same *k* and *σ* settings. We examined various *k* values. However, not all values of *k* are meaningful. For example, in case of datasets with 6 labels, the only meaningful values for comparison are *k*=3 and *k*=4. Note, that the cases in which *k*=1 and *k*=6, correspond to building a binary model and LP model, respectively. When *k*=2, INLAC and RAKEL obtain the same results, since the INLAC strategy produces all possible labelsets. When *k*=5, the number of possible models is too small.

## 5.4 Results

Table 7 presents the results obtained by averaging the 10-fold cross-validation experiments using SMO as the base classifier. The second and third columns specify the *k* and *σ* settings, respectively. The next set of columns presents the micro-$F_1$ performance. The last set of columns reports the Hamming Loss. Finally, the last line indicates the average ranking of the various methods. The results of the experimental study are encouraging. First of all, they indicate that RAKEL++ consistently improves RAKEL. Specifically, RAKEL++ obtains higher micro-$F_1$ values in 25 out of 32 cases. The BALANCOR strategy improves both the micro-averaged F-measure and the Hamming loss results of RAKEL++. The BALCO strategy mainly improves the micro-averaged F-measure compared to RAKEL++ while the INLAC strategy mainly obtains a lower Hamming loss. Thus, the BALANCOR strategy has the best of both worlds. Furthermore, the percentage improvement of DD-BALANCOR strategy over the BALANCOR strategy in terms of the micro-$F_1$ measure and Hamming Loss is 3% and 7%, respectively. Moreover, the percentage improvement of DD-BALANCOR strategy over the original RAKEL algorithm in terms of micro-averaged F-measure and Hamming Loss is 6% and 24%, respectively.

In absolute terms, the improvement in Hamming Loss seems to be insignificant (approximately 0.01). However, it should be noted that improving the Hamming Loss performance of any ensemble method is challenging since ensemble methods achieve excellent Hamming Loss performance to begin with. For example, the recently introduced EPCC algorithm [11] improved the Hamming loss of ECC by 0.003. The LEAD algorithm provides an absolute improvement of 0.003 over ECC in terms of Hamming Loss [49]. The original paper of

ECC [35] does not report Hamming Loss performance, but in a recent comparative study we have performed [6] we notice that ECC provides an absolute Hamming Loss improvement of 0.004 when compared to RAKEL. While each of the above comparisons has been performed using different settings and datasets mix, it still provides clear evidence that Hamming loss improvement is not easily achieved and that the improvement presented in this paper is not inferior to the improvements obtained by recently presented state-of-the-art algorithms.

Table 7: Comparative results using SMO as the base classifier.

| Dataset | k | σ | Micro F1 | | | | | | Hamming Loss | | | | | |
|---|---|---|---|---|---|---|---|---|---|---|---|---|---|---|
| | | | RAKEL | RAKEL++ | DD-BALANCOR | BALANCOR | INLAC | BALCO | RAKEL | RAKEL++ | DD-BALANCOR | BALANCOR | INLAC | BALCO |
| Delicious | 6 | 12441 | 0.1804 | 0.1961 | 0.1995 | 0.1991 | 0.1963 | 0.1962 | 0.092 | 0.0486 | 0.0331 | 0.0392 | 0.0463 | 0.0485 |
| Emotions | 3 | 7 | 0.679 | 0.685 | 0.6852 | 0.6851 | 0.6852 | 0.685 | 0.189 | 0.187 | 0.171 | 0.1845 | 0.1774 | 0.1865 |
| | 4 | 3 | 0.716 | 0.716 | 0.721 | 0.719 | 0.718 | 0.714 | 0.191 | 0.184 | 0.174 | 0.1755 | 0.179 | 0.1837 |
| Enron | 3 | 496 | 0.5307 | 0.5371 | 0.5476 | 0.5392 | 0.5376 | 0.5385 | 0.0553 | 0.0546 | 0.0546 | 0.0546 | 0.0546 | 0.0546 |
| | 4 | 253 | 0.5377 | 0.5394 | 0.5455 | 0.5412 | 0.5415 | 0.5392 | 0.054 | 0.0544 | 0.0541 | 0.0544 | 0.0544 | 0.0544 |
| | 5 | 213 | 0.5443 | 0.5495 | 0.5649 | 0.5543 | 0.556 | 0.5483 | 0.0527 | 0.0511 | 0.0495 | 0.0506 | 0.0508 | 0.0511 |
| Genbase | 3 | 125 | 0.981 | 0.981 | 0.9922 | 0.9888 | 0.9842 | 0.982 | 0.003 | 0.001 | 0.0007 | 0.0009 | 0.0008 | 0.001 |
| | 4 | 66 | 0.9918 | 0.9918 | 0.9922 | 0.992 | 0.992 | 0.9918 | 0.002 | 0.001 | 0.0007 | 0.001 | 0.0008 | 0.001 |
| | 5 | 44 | 0.9918 | 0.9918 | 0.9922 | 0.9921 | 0.9918 | 0.9919 | 0.003 | 0.001 | 0.0007 | 0.0008 | 0.0008 | 0.001 |
| | 6 | 31 | 0.9922 | 0.9922 | 0.9922 | 0.9922 | 0.9922 | 0.9922 | 0.003 | 0.001 | 0.0007 | 0.0008 | 0.0009 | 0.001 |
| | 7 | 22 | 0.988 | 0.988 | 0.9922 | 0.9882 | 0.9882 | 0.988 | 0.002 | 0.001 | 0.0007 | 0.0009 | 0.0009 | 0.001 |
| MediaMil | 5 | 1187 | 0.551 | 0.553 | 0.5604 | 0.5544 | 0.5544 | 0.5532 | 0.023 | 0.019 | 0.015 | 0.0173 | 0.0163 | 0.0189 |
| Medical | 3 | 372 | 0.7944 | 0.8039 | 0.8173 | 0.8081 | 0.8039 | 0.8064 | 0.0108 | 0.0079 | 0.0059 | 0.0065 | 0.0065 | 0.0078 |
| | 4 | 183 | 0.7937 | 0.8023 | 0.8156 | 0.806 | 0.8031 | 0.8044 | 0.0102 | 0.0104 | 0.0103 | 0.0104 | 0.0104 | 0.0104 |
| | 5 | 121 | 0.853 | 0.858 | 0.858 | 0.858 | 0.858 | 0.858 | 0.0112 | 0.007 | 0.0059 | 0.0066 | 0.0069 | 0.007 |
| Ohsumed | 3 | 95 | 0.3147 | 0.3238 | 0.3365 | 0.3352 | 0.3295 | 0.3252 | 0.0432 | 0.0429 | 0.0429 | 0.0429 | 0.0429 | 0.0429 |
| | 4 | 48 | 0.3138 | 0.3146 | 0.3328 | 0.3322 | 0.3233 | 0.3163 | 0.0433 | 0.0421 | 0.0411 | 0.0412 | 0.0415 | 0.042 |
| | 5 | 28 | 0.3118 | 0.3199 | 0.3398 | 0.3306 | 0.3306 | 0.3201 | 0.0441 | 0.0411 | 0.0398 | 0.0401 | 0.0407 | 0.041 |
| | 6 | 23 | 0.3076 | 0.3138 | 0.3305 | 0.3233 | 0.3233 | 0.3192 | 0.0446 | 0.0428 | 0.0421 | 0.0424 | 0.0423 | 0.0428 |
| | 7 | 17 | 0.301 | 0.3053 | 0.3276 | 0.3247 | 0.3114 | 0.3185 | 0.0459 | 0.039 | 0.0227 | 0.0304 | 0.0344 | 0.0387 |
| Scene | 3 | 7 | 0.7288 | 0.7288 | 0.7288 | 0.7288 | 0.7288 | 0.7288 | 0.0959 | 0.0914 | 0.0761 | 0.0789 | 0.087 | 0.0907 |
| | 4 | 3 | 0.7303 | 0.7304 | 0.7347 | 0.7335 | 0.7313 | 0.7306 | 0.0998 | 0.0927 | 0.0798 | 0.0891 | 0.0921 | 0.0927 |
| Slashdot | 3 | 88 | 0.7378 | 0.7379 | 0.7523 | 0.747 | 0.7387 | 0.7441 | 0.0311 | 0.0327 | 0.0299 | 0.0301 | 0.0319 | 0.0326 |
| | 4 | 45 | 0.7411 | 0.7412 | 0.7512 | 0.7484 | 0.7454 | 0.742 | 0.0287 | 0.0281 | 0.0284 | 0.0283 | 0.0283 | 0.0281 |
| | 5 | 27 | 0.7417 | 0.7435 | 0.7561 | 0.7539 | 0.7491 | 0.7465 | 0.0265 | 0.0264 | 0.0211 | 0.0242 | 0.0249 | 0.0263 |
| | 6 | 21 | 0.7489 | 0.7496 | 0.7638 | 0.7585 | 0.7498 | 0.7505 | 0.024 | 0.0235 | 0.0233 | 0.0233 | 0.0234 | 0.0235 |
| | 7 | 14 | 0.7467 | 0.7474 | 0.7688 | 0.7581 | 0.7552 | 0.7483 | 0.0331 | 0.0312 | 0.0301 | 0.031 | 0.0312 | 0.0312 |
| Yeast | 3 | 35 | 0.663 | 0.668 | 0.6698 | 0.6683 | 0.6684 | 0.6679 | 0.211 | 0.199 | 0.186 | 0.1972 | 0.1952 | 0.1985 |
| | 4 | 18 | 0.6934 | 0.6948 | 0.6963 | 0.6961 | 0.6961 | 0.6952 | 0.1968 | 0.1859 | 0.1766 | 0.1782 | 0.1851 | 0.1857 |
| | 5 | 12 | 0.6608 | 0.6628 | 0.6674 | 0.6638 | 0.6638 | 0.6628 | 0.199 | 0.1888 | 0.1773 | 0.1784 | 0.1857 | 0.1883 |
| | 6 | 9 | 0.654 | 0.661 | 0.6638 | 0.662 | 0.6622 | 0.6609 | 0.195 | 0.193 | 0.1762 | 0.19 | 0.1916 | 0.1924 |
| | 7 | 7 | 0.657 | 0.66 | 0.6622 | 0.6619 | 0.6619 | 0.6604 | 0.197 | 0.189 | 0.1795 | 0.1873 | 0.1867 | 0.189 |
| **Rank** | | | **5.2** | **4.83** | **1.39** | **2.52** | **3.02** | **4.05** | **5.59** | **4.88** | **1.19** | **2.44** | **2.91** | **4** |

The results that are obtained when the C4.5 (using WEKA's J48 implementation) is used as the base classifier are given in Table 8. As in Table 7, we notice a similar dominance of the DD-BALANCOR strategy over all other methods. Specifically, the DD-BALANCOR strategy improves both the micro-averaged F-measure and the Hamming loss results of RAKEL by 18% and 1.3%, respectively. The BALANCOR and INLAC strategies obtained a more moderate improvement. It should be noted that while in most of the datasets the new methods obtain better results than RAKEL, in the Genbase dataset none of the proposed methods offer any improvement. This might be explained by the fact that the original RAKEL algorithm has already achieved excellent results in this dataset.

The micro-averaged F-measure results in Table 7 and Table 8 indicate that the DD-BALANCOR strategy outperforms RAKEL in 57 out of 64 cases -- mostly for relatively small values of $k$ (compared to the number of labels in the dataset). On the other hand, RAKEL has not outperformed DD-BALANCOR strategy in none of the cases.

**Table 8**: Comparative results using J48 as the base classifier.

| Dataset | k | σ | Micro F1 | | | | | | Hamming Loss | | | | | |
|---|---|---|---|---|---|---|---|---|---|---|---|---|---|---|
| | | | RAKEL | RAKEL++ | DD-BALANCOR | BALANCOR | INLAC | BALCO | RAKEL | RAKEL++ | DD-BALANCOR | BALANCOR | INLAC | BALCO |
| Delicious | 6 | 12441 | 0.209 | 0.212 | 0.215 | 0.2137 | 0.212 | 0.212 | 0.0912 | 0.0626 | 0.02 | 0.0352 | 0.0611 | 0.0611 |
| Emotions | 3 | 7 | 0.6159 | 0.644 | 0.6742 | 0.6601 | 0.6482 | 0.6453 | 0.233 | 0.2232 | 0.2092 | 0.2101 | 0.2165 | 0.2234 |
| | 4 | 3 | 0.6102 | 0.6077 | 0.6138 | 0.6107 | 0.6085 | 0.6082 | 0.2563 | 0.2313 | 0.2266 | 0.2298 | 0.2296 | 0.2413 |
| Enron | 3 | 496 | 0.544 | 0.554 | 0.5602 | 0.5563 | 0.5551 | 0.5544 | 0.049 | 0.041 | 0.0407 | 0.041 | 0.0409 | 0.041 |
| | 4 | 253 | 0.553 | 0.567 | 0.581 | 0.5685 | 0.5688 | 0.5679 | 0.048 | 0.043 | 0.041 | 0.0425 | 0.0418 | 0.0427 |
| | 5 | 213 | 0.555 | 0.571 | 0.587 | 0.5804 | 0.5715 | 0.5718 | 0.047 | 0.042 | 0.0404 | 0.0415 | 0.0413 | 0.0419 |
| Genbase | 3 | 125 | 0.988 | 0.988 | 0.988 | 0.988 | 0.988 | 0.988 | 0.001 | 0.001 | 0.001 | 0.001 | 0.001 | 0.001 |
| | 4 | 66 | 0.988 | 0.988 | 0.988 | 0.988 | 0.988 | 0.988 | 0.001 | 0.001 | 0.001 | 0.001 | 0.001 | 0.001 |
| | 5 | 44 | 0.988 | 0.988 | 0.988 | 0.988 | 0.988 | 0.988 | 0.001 | 0.001 | 0.001 | 0.001 | 0.001 | 0.001 |
| | 6 | 31 | 0.988 | 0.988 | 0.988 | 0.988 | 0.988 | 0.988 | 0.001 | 0.001 | 0.001 | 0.001 | 0.001 | 0.001 |
| | 7 | 22 | 0.988 | 0.988 | 0.988 | 0.988 | 0.988 | 0.988 | 0.001 | 0.001 | 0.001 | 0.001 | 0.001 | 0.001 |
| MediaMil | 5 | 1187 | 0.551 | 0.553 | 0.5604 | 0.5531 | 0.5568 | 0.55 | 0.033 | 0.03 | 0.025 | 0.0292 | 0.0264 | 0.032 |
| Medical | 3 | 372 | 0.817 | 0.815 | 0.843 | 0.8279 | 0.8398 | 0.826 | 0.011 | 0.009 | 0.008 | 0.0089 | 0.0088 | 0.0089 |
| | 4 | 183 | 0.787 | 0.788 | 0.7953 | 0.7917 | 0.7902 | 0.7887 | 0.013 | 0.011 | 0.011 | 0.0111 | 0.0111 | 0.011 |
| | 5 | 121 | 0.821 | 0.819 | 0.864 | 0.8287 | 0.8203 | 0.8204 | 0.009 | 0.009 | 0.009 | 0.009 | 0.009 | 0.009 |
| Ohsumed | 3 | 95 | 0.276 | 0.3731 | 0.485 | 0.3914 | 0.408 | 0.3743 | 0.065 | 0.063 | 0.0622 | 0.0627 | 0.0626 | 0.063 |
| | 4 | 48 | 0.285 | 0.331 | 0.511 | 0.3465 | 0.3887 | 0.3364 | 0.065 | 0.062 | 0.0591 | 0.059 | 0.0618 | 0.0617 |
| | 5 | 28 | 0.286 | 0.362 | 0.5172 | 0.3752 | 0.4152 | 0.3638 | 0.066 | 0.0607 | 0.0584 | 0.06 | 0.0603 | 0.0602 |
| | 6 | 23 | 0.281 | 0.2995 | 0.476 | 0.314 | 0.3288 | 0.3092 | 0.066 | 0.0602 | 0.0591 | 0.0601 | 0.0594 | 0.06 |
| | 7 | 17 | 0.269 | 0.317 | 0.4 | 0.3323 | 0.3236 | 0.306 | 0.067 | 0.0603 | 0.0594 | 0.0595 | 0.0599 | 0.0602 |
| Scene | 3 | 7 | 0.6798 | 0.6804 | 0.7044 | 0.6917 | 0.683 | 0.6814 | 0.1145 | 0.1068 | 0.0982 | 0.1019 | 0.1014 | 0.106 |
| | 4 | 3 | 0.6327 | 0.638 | 0.672 | 0.6472 | 0.6413 | 0.6388 | 0.1476 | 0.1196 | 0.1062 | 0.1076 | 0.1116 | 0.1195 |
| Slashdot | 3 | 88 | 0.338 | 0.3834 | 0.4814 | 0.3988 | 0.4124 | 0.3838 | 0.043 | 0.043 | 0.041 | 0.0414 | 0.0421 | 0.0431 |
| | 4 | 45 | 0.342 | 0.4011 | 0.4974 | 0.4234 | 0.4183 | 0.4029 | 0.044 | 0.043 | 0.042 | 0.0426 | 0.0427 | 0.0429 |
| | 5 | 27 | 0.355 | 0.3923 | 0.4666 | 0.4196 | 0.4101 | 0.395 | 0.045 | 0.042 | 0.042 | 0.042 | 0.042 | 0.042 |
| | 6 | 21 | 0.353 | 0.3849 | 0.4392 | 0.4081 | 0.4033 | 0.389 | 0.044 | 0.042 | 0.042 | 0.042 | 0.042 | 0.042 |
| | 7 | 14 | 0.336 | 0.347 | 0.408 | 0.3696 | 0.3576 | 0.351 | 0.046 | 0.043 | 0.043 | 0.043 | 0.043 | 0.043 |
| Yeast | 3 | 35 | 0.596 | 0.592 | 0.611 | 0.6001 | 0.5981 | 0.5939 | 0.242 | 0.236 | 0.219 | 0.2167 | 0.2341 | 0.2362 |
| | 4 | 18 | 0.591 | 0.588 | 0.623 | 0.6002 | 0.59 | 0.5903 | 0.236 | 0.228 | 0.218 | 0.2175 | 0.2229 | 0.228 |
| | 5 | 12 | 0.601 | 0.6088 | 0.644 | 0.627 | 0.619 | 0.6122 | 0.245 | 0.216 | 0.219 | 0.2163 | 0.2178 | 0.2161 |
| | 6 | 9 | 0.62 | 0.6279 | 0.651 | 0.6338 | 0.6315 | 0.63 | 0.233 | 0.214 | 0.216 | 0.2151 | 0.2144 | 0.2144 |
| | 7 | 7 | 0.614 | 0.634 | 0.643 | 0.6347 | 0.6363 | 0.624 | 0.221 | 0.213 | 0.2081 | 0.211 | 0.2128 | 0.2123 |
| Rank | | | 5.67 | 4.77 | 1.16 | 2.41 | 2.94 | 4.06 | 5.52 | 4.08 | 2.06 | 2.8 | 2.97 | 3.64 |

Following the procedure presented by Demsar [12], we compared the various algorithms according to their average rank. Noticeably, the DD-BALANCOR strategy obtained the highest average rank both in the micro-$F_1$ and the Hamming Loss cases. The BALANCOR strategy achieved the second best average rank. The null hypothesis that all methods have the same micro-$F_1$ measure performance was rejected by the adjusted Friedman test with a confidence level of 95% and (5, 315) degrees of freedom (specifically F(5,315)=139.46 > 2.24 and p-value<0.0001).

As the null hypothesis was rejected, we proceeded to the Nemenyi post-hoc test. In this case two classifiers are *significantly different* with a confidence level of 95% if their average ranks differ by at least 0.942. Thus, we observe that DD-BALANCOR significantly outperforms all other methods. However, we could not reject the null hypothesis that RAKEL and RAKEL++ have the same micro F-Measure performance at confidence levels of 95%. Additionally, we could not reject the null hypothesis that BALANCOR and INLAC have the same micro F-Measure.

In case of the Hamming loss, the null hypothesis that all methods have the same Hamming loss performance was rejected by the adjusted Friedman test with a confidence level of 95% and (5, 315) degrees of freedom (specifically, F(5,315)=86.15 > 2.24 and p-value<0.0001). Using the Nemenyi post-hoc test we concluded that DD-BALANCOR significantly outperforms all other methods in terms of Hamming loss. Moreover RAKEL++ significantly outperformed RAKEL.

## 5.5 Comparison to Other Ensembles Methods

Table 9 compares the predictive performance of the DD- BALANCOR with two state-of-the-art ensemble algorithms: ECC (Ensembles of Classifier Chains) and Ensembles of Pruned Sets (EPS). These two algorithms were chosen due to their popularity and their availability in the MULAN package. To reduce the computational cost of the experiments, we tested these algorithms using only J48 as the base classifier. The ensemble size of ECC and EPS algorithms was adjusted to that of the DD-BALANCOR algorithm. Specifically, in the case of ECC, the ensemble size is an integer product of the number of labels. Thus, we selected the smallest ensemble size which is greater or equal to the ensemble size obtained by DD-BALANCOR.

The results presented in Table 9 are very encouraging. It can be seen that DD-BALANCOR succeeded to improve the performance of RAKEL to the level of ECC. In particular, in 7 out of 10 cases, DD-BALANCOR obtained the lowest Hamming Loss while in the remaining datasets the performance of DD-BALANCOR was only slightly worse than the performance of ECC.

We compared the various algorithms according to their average rank. Noticeably, the DD-BALANCOR strategy obtained the highest average rank in both the micro-$F_1$ and the Hamming Loss measures. The ECC algorithm achieved the second best average rank. The null hypothesis that all methods have the same Hamming Loss performance was rejected by the adjusted Friedman test with a confidence level of 95% and (2, 20) degrees of freedom (specifically F(2,20)= 53.068 > 3.492 and p-value<0.0001). Since the null hypothesis was rejected and since we wanted to examine whether our newly proposed method is better than existing ones, following Demsar [12] we used Bonferroni correction post-hoc test instead of Nemenyi post-hoc test that was used in the previous section. Using the Bonferroni test, we found that DD-BALANCOR significantly outperforms ECC and EPS with p<9% and p<1%, respectively.

The last two columns in Table 9 compare the training time of the algorithms. Note that the training time of DD-BALANCOR includes only the time required for deriving the best permutation together with the training of the ensemble of classifiers. It does not include the time required for the matrix construction as we assume that this task is performed off-line only once and in advance. As we can see, the training time of ECC and DD-BALANCOR are mostly the same order of magnitude. However, the training time of EPS is consistently lower than that of the two other methods. The last column in Table 9 presents the computational cost required for constructing the BALANCOR matrix. It can be seen that the matrix construction time is not negligible for large matrices. This emphasizes the need to separate the matrix construction from the ensemble training.

Table 9: Comparing DD-BALANCOR with ECC using J48 as the base classifier.

| Dataset | Ensemble Size σ | micro-averaged F-measure | | | Hamming loss | | | Training Time (in seconds) | | | BALANCOR Matrix Designing Time |
|---|---|---|---|---|---|---|---|---|---|---|---|
| | | DD-BALANCOR | ECC | EPS | DD-BALANCOR | ECC | EPS | DD-BALANCOR | ECC | EPS | |
| Delicious | 12441 | 0.215 | 0.199 | 0.187 | 0.02 | 0.09295 | 0.06628 | 153769.8 | 76224.74 | 44163.86 | 94776.3 |
| Emotions | 7 | 0.6742 | 0.6684 | 0.6678 | 0.2092 | 0.2124 | 0.2162 | 2.83 | 2.23 | 1.453 | 0.055042 |
| Enron | 496 | 0.5602 | 0.5451 | 0.4792 | 0.0407 | 0.0394 | 0.0494 | 3559.22 | 2587.06 | 65.263 | 6604.55 |
| Genbase | 66 | 0.988 | 0.9877 | 0.9839 | 0.001 | 0.001 | 0.0015 | 12.12 | 16.67 | 1.465 | 213.37 |
| MediaMill | 1187 | 0.5604 | 0.5621 | 0.5619 | 0.025 | 0.031 | 0.0318 | 63198.29 | 180737 | 39759 | 37314.23 |
| Medical | 183 | 0.7953 | 0.7866 | 0.7649 | 0.011 | 0.0112 | 0.0126 | 255.57 | 240.66 | 9.643 | 12636.84 |
| Ohsumed | 48 | 0.511 | 0.5172 | 0.4439 | 0.0591 | 0.059 | 0.0612 | 11650.31 | 12138.95 | 674.346 | 50.44 |
| Scene | 7 | 0.7044 | 0.6839 | 0.684 | 0.0982 | 0.1035 | 0.1076 | 31.52 | 38.42 | 21.056 | 0.055042 |
| Slashdot | 88 | 0.4814 | 0.4825 | 0.4667 | 0.041 | 0.0425 | 0.0473 | 1792.8 | 2210.63 | 514.931 | 11.34 |
| Yeast | 7 | 0.643 | 0.6291 | 0.6178 | 0.2081 | 0.215 | 0.2276 | 68.82 | 46.81 | 30.053 | 0.055042 |
| Rank | | 1.4 | 1.8 | 2.8 | 1.3 | 1.9 | 2.9 | 2.5 | 2.5 | 1 | |

## 5.6 Using higher power values

As Lemma 2 indicates, using higher values of $r$, has a direct impact on the matrix size and consequently on the computational cost required for training the ensemble. Thus, using matrices based on $r = 3$ is practical only for datasets with a relatively small label set. In Table 10 we present the predictive performance obtained by the DD-BALANCOR algorithm for $r = 3$ and compare it to the performance obtained for $r = 2$. In addition we present the performance of RAKEL++ using the same ensemble sizes of $r = 2$ and $r = 3$. Note that we have tested $r = 3$ only on small datasets since the training time for larger datasets could not be completed within one week given the resources that were used for the experiments. The results in Table 10 indicate that using higher values of $r$ improve the classification accuracy. One can notice that RAKEL++ improves the Hamming Loss by 7% on average. Similarly, the average improvement of DD-BALANCOR is approximately 9%. Thus, the DD-BALANCOR algorithm is still better than its equivalent RAKEL++ for $r = 3$. Moreover, one can see that by increasing the ensemble size of RAKEL++ (equivalent to $r = 3$), we succeeded to outperform the performance of DD-BALANCOR with $r = 2$. Thus, if computational cost is not a concern, one can use random strategy and simply increase the ensemble size in order to improve the predictive performance until it reaches an asymptotic value. However, if the computational cost is important, strategies such as DD-BALANCOR can assist in finding a compact ensemble.

Table 10: Comparative results for $r = 2$ and $r = 3$ using J48 as the base classifier.

| Dataset | k | σ – Ensemble Size | | Micro F1 | | | | Hamming Loss | | | |
|---|---|---|---|---|---|---|---|---|---|---|---|
| | | DD-BALANCOR with r=2 | DD-BALANCOR with r=3 | DD-BALANCOR with r=2 | DD-BALANCOR with r=3 | RAKEL++ equivalent to r=2 | RAKEL++ equivalent to r=3 | DD-BALANCOR with r=2 | DD-BALANCOR with r=3 | RAKEL++ equivalent to r=2 | RAKEL++ equivalent to r=3 |
| Emotions | 4 | 3 | 6 | 0.6138 | 0.6296 | 0.6077 | 0.6280 | 0.2266 | 0.2132 | 0.2313 | 0.2260 |
| Ohsumed | 4 | 48 | 443 | 0.5110 | 0.7088 | 0.3310 | 0.5657 | 0.0591 | 0.0552 | 0.0620 | 0.0562 |
| | 5 | 28 | 178 | 0.5172 | 0.7452 | 0.3620 | 0.5944 | 0.0584 | 0.0566 | 0.0607 | 0.0581 |
| | 6 | 23 | 89 | 0.4760 | 0.7517 | 0.2995 | 0.6101 | 0.0591 | 0.0577 | 0.0602 | 0.0580 |
| | 7 | 17 | 51 | 0.4000 | 0.4963 | 0.3170 | 0.4252 | 0.0594 | 0.0585 | 0.0603 | 0.0588 |
| Scene | 4 | 3 | 6 | 0.6720 | 0.6823 | 0.6380 | 0.6740 | 0.1062 | 0.0896 | 0.1196 | 0.1012 |
| Slashdot | 4 | 45 | 386 | 0.4974 | 0.6223 | 0.4011 | 0.5359 | 0.0420 | 0.0409 | 0.0430 | 0.0420 |
| | 5 | 27 | 155 | 0.4666 | 0.5228 | 0.3923 | 0.4811 | 0.0420 | 0.0320 | 0.0420 | 0.0370 |
| | 6 | 21 | 78 | 0.4392 | 0.5538 | 0.3849 | 0.4918 | 0.0420 | 0.0320 | 0.0420 | 0.0380 |
| | 7 | 14 | 45 | 0.4080 | 0.4568 | 0.3470 | 0.4343 | 0.0430 | 0.0330 | 0.0430 | 0.0360 |
| Yeast | 4 | 18 | 92 | 0.6230 | 0.6684 | 0.5880 | 0.6340 | 0.2180 | 0.1893 | 0.2280 | 0.1990 |
| | 5 | 12 | 37 | 0.6440 | 0.6358 | 0.6088 | 0.6270 | 0.2190 | 0.2041 | 0.2160 | 0.2030 |
| | 6 | 9 | 19 | 0.6510 | 0.6302 | 0.6279 | 0.6170 | 0.2160 | 0.2124 | 0.2140 | 0.2120 |
| | 7 | 7 | 11 | 0.6430 | 0.6061 | 0.6340 | 0.6000 | 0.2081 | 0.1929 | 0.2130 | 0.2020 |

## 5.7 Discussion

The proposed strategies perform in a highly efficient and stable manner where in many cases they obtain better results than RAKEL, especially for small values of $k$ relatively to the number of labels. These results are consistent with Lemma 1 and Lemma 2. The ensemble size $\sigma$ has a direct impact on the training time cost as it indicates the number of base-classifiers that are trained. Thus, if the parameter $k$ is kept fixed, increasing the value of $\sigma$ will usually increase the training time linearly. On the other hand, it is well known that the parameter $k$ allows a trade-off between predictive performance and training time costs [35]. To some extent, using higher values of $k$ for the same ensemble size $\sigma$ increases the training time cost but also improves the predictive performance. In this context, a property that is shared by all four proposed construction strategies is that increasing the subset size $k$ results in a decrease in the ensemble size $\sigma$ required to ensure coverage. The number of classes addressed by each ensemble member is bounded by $min\{2^k, |T|\}$ where $|T|$ is the training set size. Practically, the actual number of classes depends on the characteristics of the dataset in hand (such as the Label Density).

As the actual number of classes approaches the upper bound, it might negatively affect the learning procedure in two aspects: First, the training time increases because, at the very least, it is linear in the number of classes. Second, the number of training instances associated with each class is relatively small which makes it harder for the base-learning algorithm to differentiate among the classes. In fact, within the PAC framework, Ben-David et al. [3] showed how the required sample size depends on the dimensionality of the classes. Clearly, every ensemble should be large enough to include a sufficient number of votes for each label. This insight might explain the clear difference in results for the smallest datasets (in terms of labels) for *k=3* and *k=4*. On the other hand, for datasets with a higher number of labels such as Slashdot, the number of models is large enough and thus there is no such difference in the results.

Given all readymade matrices that were constructed using the abovementioned strategies, the user can choose the matrix (and consequently also the value of $k$ and $r$) according to various criteria. From the computational cost perspective, one can estimate the cost based on the number of rows in the matrix (ensemble size), the associated $k$ value, and how the computational complexity of the intended base inducer is affected by $k$ and the training set size (number of instances and number of attributes). From the accuracy point of view, one can use the dataset characteristics, such as label cardinality, as a hint for selecting the matrix. It can be presumably claimed that higher values of $k$ should be preferred when the label cardinality is high. However, recall that higher values of $k$ result in smaller ensemble sizes which in turn can potentially decrease the predictive performance. The experimental results in Tables 7 and 8 show that higher values of $k$ do not necessarily result in better performance. Thus, we suggest using the data driven objective function presented in Section 4.4 also as a means for choosing the matrix. Namely, from all available matrices that fit the current problem, one should choose the matrix which maximizes the total dependency that is addressed by the matrix. We put this idea to the test. For each dataset[4] in Tables 7 and 8 we tested if the suggested procedure could correctly choose between the lowest and the highest value of $k$ (i.e. binary selection). The selection is considered to be correct if it selects the matrix configuration with the lowest Hamming Loss. The suggested procedure selected the correct matrix in 64 out of 80 cases in Table 7 (8 datasets times 10 folds) and 59 out of 80 cases in Table 8. This indicates a mean accuracy of 75%. While this is not perfect, it is still much better than a random guess.

As for the value of $r$, we showed analytically (Lemma 2) and experimentally (see Section 5.4) that higher values of $r$ increase sustainably the ensemble size. Nonetheless, using higher values of $r$ improves the predictive performance as demonstrated in Section 5.4. Since computational cost is usually a constraint, we recommend determining the values of $r$ and $k$ using the lower bound of Lemma 2 multiplied by the base inducer computational cost for inducing a single classifier. The product should not exceed the given constraint.

From the practitioner point of view, given RAKEL++, using BALANCOR requires neither any additional learning time nor any significant revisions to the RAKEL++ code. Simply instead of using random label-set, one should use the predefined label-set matrices. The Matlab code for generating the matrices can be obtained from http://www.ise.bgu.ac.il/faculty/liorr/matlab.rar.

## 6. Conclusions and Future Work

In this paper we presented a new ensemble method for multi-label classification. We examined the hypothesis that constructing compact ensembles which obey certain constraints may achieve predictive results that are better than the results obtained by RAKEL. Namely, find a minimal number of label-sets of predefined size for a given number of labels and a set of constraints. We introduced the set cover problem as a general framework for constructing such ensembles and demonstrated its effectiveness when the following constraints were enforced: (a) equal representation of each individual label; (b) coverage of inter-label correlations; and (c) combination of (a) and (b). Finally, we presented a data-driven method which adjusts the matrix according to the dataset in hand.

Our experimental study shows that the method performs in a highly efficient and stable manner in many cases better than RAKEL and other ensemble methods. The proposed method resolves two main problems of the RAKEL algorithm: (1) random selection of label-sets that may produce fluctuations in the result values; and (2) the need to define the number of models in an ensemble.

---

[4] Except for Delicous, MediaMill which due to their computational intensive cost we were able to test only one configuration

The ability to enforce constraints during the cover construction along with the results achieved in this paper motivate the investigation of other constraints which may produce more powerful coverage schemes that will further improve the predictive performance.

## References

[1] Allwein, E. L., Schapire, R. E., & Singer, Y. (2000). Reducing Multiclass to Binary: A Unifying Approach for Margin Classifiers, JMLR 1:113–141.
[2] Kisilevich S., Rokach L, Y Elovici, B Shapira (2010), Efficient multidimensional suppression for k-anonymity, IEEE Transactions on Knowledge and Data Engineering, 22(3):334-347.
[3] Ben-David, S. and Cesabianchi, N. and Haussler, D. and Long, P.M. (1995), Characterizations of Learnability for Classes of (0,...,n)-Valued Functions.
[4] Boutell, M., Luo, J., Shen, X., & Brown, C. (2004). Learning multi-label scene classification. Pattern Recognition, Vol. 37(9), pp. 1757–1771.
[5] Brinker, K., Furnkranz, J., & Hullermeier, E. (2006). A unified model for multilabel classification and ranking. Proceedings of the 17th European Conference on AI.
[6] Chekina L., Rokach L., and Shapira B. (2011), Meta-Learning for Selecting a Multi-Label Classification Algorithm, IEEE 11th International Conference on Data Mining Workshops (ICDMW), pp. 220-227.
[7] Chvatal, V. (1979). A greedy heuristic for the set-covering problem, Mathematics of Operations Research, 4, 233–235.
[8] Clare, A., & King, R. D. (2001). Knowledge Discovery in Multi-label Phenotype Data. Lecture Notes in Computer Science, Vol. 2168, Springer, Berlin.
[9] Cochran, W. G. (1954). Some methods for strengthening the common ?2 tests. Biometrics, 10, pp. 417–451.
[10] Dawson, D.A., & Sankoff, D. (1967). An inequality for probabilities. Proceedings of the American Mathematical Society, 18, pp. 504–507.
[11] Dembczynski, K., Cheng, W., & Hüllermeier, E. (2010). Bayes Optimal Multilabel Classification via Probabilistic Classifier Chains. ICML.
[12] Demsar, J. (2006). Statistical Comparisons of Classifiers over Multiple Data Sets. Journal of Machine Learning Research, Vol. 7, pp. 1–30.
[13] Dietterich, T. G., & Bakiri, G. (1995). Solving Multiclass Learning Problems via Error-Correcting Output Codes. Journal of AI Research, Vol. 2, pp. 263–286.
[14] Enron Email Analysis Project. http://bailando.sims.berkeley.edu/enron_email.html. UC Berkeley.
[15] Fan R.-E. and Lin C.-J., A study on threshold selection for multi-label classification. Department of Computer Science, National Taiwan University, 2007.
[16] Frank, E., Hall, M. A., Holmes, G., Kirkby, R., Pfahringer, B., & Witten, I. H., (2005). Weka: A machine learning workbench for data mining.
[17] Garey, M. R., & Johnson, D. S. (1979). Computers and Intractability — A Guide to the Theory of NP-Completeness (W. H. Freeman and Company).
[18] Garfinkel, R. S., & Nemhauser, G. L. (1972). Integer Programming (John Wiley & Sons).
[19] Ghani, R., (2000). Using Error-Correcting Codes for Text Classification. Proceedings of the 17th International Conference on Machine Learning, pp. 303–310.
[20] Gomes, F. C., Meneses, C. N., Pardalos, P. M., & Viana, G. V. R. (2006). Experimental analysis of approximation algorithms for the vertex cover and set covering problems.
[21] Grossman, T., & Wool., A. (1997). Computational experience with approximation algorithms for the set covering problem, European Journal of Operational Research.
[22] Har-Peled, S., Roth, D., & Zimak, D. (2002). Constraint Classification: A New Approach to Multiclass Classification and Ranking, Algorithmic Learning Theory, pp. 267–280.
[23] Hedayat, A. S., Sloane, N. J. A., & Stufken, J. (1999). Orthogonal Arrays: Theory and Applications. Springer-Verlag, New York.
[24] Itach, E., Tenenboim, L., & Rokach, L. (2009). An Ensemble Method for Multi-label Classification using a Transportation Model, ECML Workshop.
[25] Ji, L., & Yin, J., (2010). Constructions of new orthogonal arrays and covering arrays of strength three, Journal of Combinatorial Theory, Series A 117, pp. 236–247.

[26] Karp, R. M. (1972), Reducibility among combinatorial problems, Complexity of computer computations (R.E. Miller and J.W. Thatcher, eds.), Plenum Press, New-York, 85–103.
[27] Kiefer, J. (1953). Sequential minimax search for a maximum, Proceedings of the American Mathematical Society, Vol. 4 (3), pp. 502–506.
[28] Kohavi, R., & John, G.-H. (1997). Wrappers for Feature Subset Selection. Artifitial Intelligence 97, Vol. (1-2), pp. 273–324.
[29] Lau, H. T. (2007). A Java Library of Graph Algorithms and Optimization. Taylor & Francis Group.
[30] Niculescu-Mizil, A., & Caruana, R. (2005). Obtaining Calibrated Probabilities from Boosting.
[31] Östergård, P. R. J. (2010). Classification of binary constant weight codes. IEEE Transactions on Information Theory, 56(8): 3779–3785.
[32] Park, S. H., & Fürnkranz, J. (2011). Efficient prediction algorithms for binary decomposition techniques, Data Mining and Knowledge Discovery.
[33] Peleg, D., Schechtman, G., & Wool, A. (1997). Randomized approximation of bounded multicovering problems. Algorithmica, 18(1), pp. 44–66.
[34] Read, J., Pfahringer, B., & Holmes, G. (2008). Multi-label Classification Using Ensembles of Pruned Sets. International Conference of Data Mining, pp. 995–1000, Pisa, Italy.
[35] Read, J., Pfahringer B., Holmes, J., & Frank, E. (2009). Classifier Chains for Multi-label Classification.
[36] Rokach, L., & Itach, E. (2010). An Ensemble Method for Multi-label Classification using an Approximation Algorithm for the Set Covering Problem.
[37] Matatov N., Rokach L., Maimon O. (2010), Privacy-preserving data mining: A feature set partitioning approach, Information Sciences 180(14):2696-2720.
[38] Snoek, C.G., Worring, M., van Gemert, J.C., Geusebroek, J.M., & Smeulders, A.W. (2006). The challenge problem for automated detection of 101 semantic concepts in mul-timedia.
[39] Chekina, L. and Gutfreund, D. and Kontorovich, A. and Rokach, L. and Shapira, B. (2013), Exploiting label dependencies for improved sample complexity, Machine Learning, 91(1):1-42.
[40] Trohidis, K., Tsoumakas, G., Kalliris, G., & Vlahavas, I. (2008). Multilabel Classification of Music into Emotions.
[41] Tsoumakas, G., & Katakis, I. (2007). Multi-label classification: An overview. International Journal of Data Warehousing and Mining, pp. 1–13.
[42] Tsoumakas, G., Katakis, I., & Vlahavas, I. (2008). Effective and Efficient Multilabel Classification in Domains with Large Number of Labels, Proc. ECML/PKDD 2008 Workshop.
[43] Tsoumakas, G. and Katakis, I. and Vlahavas I. (2011), Random k-labelsets for multi-label classification, IEEE Transactions on Knowledge and Data Engineering, 23(7): 1079-1089
[44] Tsoumakas G., Partalas, I., & Vlahavas, I. P. (2009). An Ensemble Pruning Primer, Applications of Supervised and Unsupervised Ensemble Methods, pp. 1–13.
[45] Tsoumakas, G., & Vlahavas, V. (2007). Random k-Labelsets: An Ensemble Method for Multilabel Classification. The 18th European Conference on Machine Learning (ECML).
[46] Wolpert, D. H. (1992). Stacked generalization, Neural Networks, Vol. 5, pp. 241–259.
[47] Yang, Y. (1999). An evaluation of statistical approaches to text categorization. Journal of Information Retrieval, Vol. 1, pp. 78–88.
[48] Zhang, A., Wu, Z.-L., Li, C.-H., & Fang, K.-T. (2003). On Hadamard-Type Output Coding in Multiclass Learning, Intelligent Data Engineering and Automated Learning, Vol. 2690,
[49] Zhang, M.-L. & Zhang, K. (2010). Multi-label learning by exploiting label dependency, 16th ACM SIGKDD Conference on Knowledge Discovery and Data mining (KDD 2010), pp. 999–10
[50] Zhou, Z.-H., Wu, J., & Tang, W. (2002). Ensembling neural networks: many could be better than all. Artificial Intelligence, 137(1-2), pp. 239–263.
[51] Zinoviev, V. A. (1996). On the equivalence of certain constant weight codes and combinatorial designs, Journal of Statistical Planning and Inference, Vol. 56(2), pp. 289–294.
[52] Menahem E., Rokach L., Elovici Y. (2009), Troika–An improved stacking schema for classification tasks, Information Sciences 179 (24):4097-4122

# Appendix

**Lemma 1**: The probability that all pairs in a label set $L = \{\lambda_i\}_{i=1}^m$ will be covered by $\sigma$ random $k$-labelsets is bounded by:

$$p \leq 1 - \frac{2}{h+1}S_1 + \frac{2}{h(h+1)}S_2$$

where:

$$S_1 = \binom{m}{2}\left(\frac{\binom{m-2}{k} + 2\binom{m-2}{k-1}}{\binom{m}{k}}\right)^\sigma = \frac{m(m-1)}{2} \cdot \left(\frac{(m-k)(m+k-1)}{m(m-1)}\right)^\sigma$$

$$S_2 = 3\binom{m}{4}\left(\frac{\binom{m-4}{k} + 4\binom{m-4}{k-1} + 4\binom{m-4}{k-2}}{\binom{m}{k}}\right)^\sigma + 3\binom{m}{3}\left(\frac{\binom{m-3}{k} + 3\binom{m-3}{k-1} + \binom{m-3}{k-2}}{\binom{m}{k}}\right)^\sigma$$

and

$$h = 1 + \left\lfloor\frac{2S_2}{S_1}\right\rfloor.$$

**Proof**

Let $e_{i,j}$ denote the event that the interaction of labels $\lambda_i$ and $\lambda_j$ (we assume $j > i$) are not covered by the cover $I$. We are interested in bounding the probability for the case which not all interactions of labels-pairs are covered. Namely we need to look into the expression $1 - Pr(\bigcup_{\forall i,j;j>i} e_{i,j})$.

Using the Dawson-Sankoff [10] inequality, the probability of the union of events is bounded by:

$$Pr(\bigcup_{\forall i,j;j>i} e_{i,j}) \geq \frac{2}{h+1}S_1 - \frac{2}{h(h+1)}S_2$$

where

$$S_1 = \sum_{\forall i,j:j>i} Pr(e_{i,j})$$

$$S_2 = \sum_{\substack{\forall i,j,s,t \\ j>i;t>s;i\neq s;j\neq t}} Pr(e_{i,j} \cap e_{s,t})$$

and

$$h = 1 + \left\lfloor\frac{2S_2}{S_1}\right\rfloor.$$

Note that $S_1$ looks into covering of individual pairs of labels while $S_2$ looks into combinations of covered pairs. In order to calculate $Pr(e_{i,j})$, we first calculate the probability that a single $k$-labelset does not cover a certain pair of labels. This is given by:

$$Pr(\lambda_i, \lambda_j) = \frac{\binom{m-2}{k} + 2\binom{m-2}{k-1}}{\binom{m}{k}}$$

Thus, the probability that none of $\sigma$ randomly chosen $k$-labelsets cover a certain pair of labels is given by:

$$Pr(e_{i,j}) = \left(\frac{\binom{m-2}{k} + 2\binom{m-2}{k-1}}{\binom{m}{k}}\right)^{\sigma}$$

Since we have $\binom{m}{2}$ pairs of labels in $S_1$, its value is:

$$S_1 = \binom{m}{2}\left(\frac{\binom{m-2}{k} + 2\binom{m-2}{k-1}}{\binom{m}{k}}\right)^{\sigma} = \frac{m(m-1)}{2} \cdot \left(\frac{(m-k)(m+k-1)}{m(m-1)}\right)^{\sigma}$$

The value of $S_2$ can be decomposed into two complementary cases:

$$S_2 = S_{2,1} + S_{2,2}$$

**Case 1:** The two pairs are mutually exclusive ($i \neq j, j \neq s, i \neq t, j \neq t$).

We first calculate the probability that a single $k$-labelset does not to cover the two pairs $(\lambda_i, \lambda_j)$ and $(\lambda_s, \lambda_t)$:

$$Pr((\lambda_i, \lambda_j) \cap (\lambda_s, \lambda_t)) = \frac{\binom{m-4}{k} + 4\binom{m-4}{k-1} + 4\binom{m-4}{k-2}}{\binom{m}{k}}$$

Note that the first element in the enumerator refers to all cases in which none of the subject labels is covered by a $k$-labelset. The second element refers to all cases in which exactly one of the labels is covered. The third element refers to all cases in which exactly two labels are covered (one from each pair). Since there are $\sigma$ $k$-labelsets we obtain:

$$Pr(e_{i,j} \cap e_{s,t}) = \left(\frac{\binom{m-4}{k} + 4\binom{m-4}{k-1} + 4\binom{m-4}{k-2}}{\binom{m}{k}}\right)^{\sigma}$$

In case 1, there are $3\binom{m}{4}$ couples of label-pairs. Note that we multiply the number of combinations by 3 since there are exactly three different ways to partition four labels into two pairs. Thus we conclude that:

$$S_{2,1} = 3\binom{m}{4}\left(\frac{\binom{m-4}{k} + 4\binom{m-4}{k-1} + 4\binom{m-4}{k-2}}{\binom{m}{k}}\right)^{\sigma}$$

**Case 2:** The two pairs are not mutually exclusive.

Using the same argument as in case 1:

$$Pr((\lambda_i, \lambda_j) \cap (\lambda_s, \lambda_t)) = \frac{\binom{m-3}{k} + 3\binom{m-3}{k-1} + \binom{m-3}{k-2}}{\binom{m}{k}}$$

Therefore:

$$S_{2,2} = 3\binom{m}{3}\left(\frac{\binom{m-3}{k} + 3\binom{m-3}{k-1} + \binom{m-3}{k-2}}{\binom{m}{k}}\right)^{\sigma} \blacksquare$$